\documentclass[AMA,STIX1COL]{WileyNJD-v2}
\usepackage{moreverb}
\usepackage{enumerate}
\usepackage{subfigure}

\newcommand\BibTeX{{\rmfamily B\kern-.05em \textsc{i\kern-.025em b}\kern-.08em
T\kern-.1667em\lower.7ex\hbox{E}\kern-.125emX}}

\articletype{Article Type}%

\received{<day> <Month>, <year>}
\revised{<day> <Month>, <year>}
\accepted{<day> <Month>, <year>}

\begin{document}

\title{Proximal Policy Optimization Learning based Control of Congested Freeway Traffic}

\author[1]{Shurong Mo}

\author[1,2]{Nailong Wu}

\author[1,2]{Jie Qi}

\author[1]{Anqi Pan}

\author[3]{Zhiguang Feng}

\author[4]{Huaicheng Yan}

\author[5]{Yueying Wang}

\authormark{Mo \textsc{et al}}

\address[1]{College of Information Science and Technology, Donghua University, Shanghai 201620, China}

\address[2]{Engineering Research Center of Digitized Textile and Fashion Technology Ministry of Education, Donghua University, Shanghai 201620, China}

\address[3]{College of Intelligent Systems Science and Engineering, Harbin Engineering University, Harbin, 150056, China}

\address[4]{School of Information Science and Engineering, East China University of Science and Technology,  Shanghai 200237, China}

\address[5]{School of Mechatronic Engineering and Automation, Shanghai University, Shanghai 200444, China}

\corres{Nailong Wu, College of Information Science and Technology, Donghua University, Shanghai 201620, China. \\ Engineering Research Center of Digitized Textile and Fashion Technology Ministry of Education, Donghua University, Shanghai 201620, China.\\ Email: nathan\_wu@dhu.edu.cn}

\presentaddress{Present address}

\abstract[Abstract]{	In this paper, a delay compensation feedback controller based on reinforcement learning is therefore proposed to adjust the time interval of the adaptive cruise control (ACC) vehicle agents in traffic congestion by introducing the proximal policy optimization (PPO) scheme. The high-speed traffic flow is characterized by two-by-two Aw Rasle Zhang (ARZ) nonlinear first-order partial differential equations (PDEs). Unlike the backstepping delay compensation control \cite{qi2021delaycompensated}, the PPO controller proposed in this paper consists of the current traffic flow velocity, the current traffic flow density, and the previous one-step control input. Since the system dynamics of the traffic flow are difficult to be expressed mathematically, the control gains of the three feedback can be determined via learning from the interaction between the PPO and the digital simulator of the traffic system. The performance of Lyapunov control, backstepping control and PPO control are compared with numerical simulation. The results demonstrate that PPO control is superior to Lyapunov control in terms of the convergence rate and control efforts for the traffic system without delay. As for the traffic system with an unstable input delay value, the performance of the PPO controller is also equivalent to that of the backstepping controller. Besides, PPO is more robust than backstepping controller when the parameter is sensitive to Gaussian noise.}



\jnlcitation{\cname{%
\author{Mo S},
\author{Wu N},
\author{Qi J},
\author{Pan A},
\author{Feng Z},
\author{Yan H}, and
\author{Wang Y}} (\cyear{2022}),
\ctitle{Proximal Policy Optimization Learning based Control of Congested Freeway Traffic}, \cvol{2022;00:1--6}.}

\maketitle

\footnotetext{\textbf{Abbreviations:} ANA, anti-nuclear antibodies; APC, antigen-presenting cells; IRF, interferon regulatory factor}

\section{Introduction}\label{sec1}
Traffic congestion is one of the most persistent problems of society today. Therefore, how to alleviate traffic congestion has always been a hot research topic in the field of traffic control, which has attracted widespread attentions \cite{dadashova2021multivariate}.
On the freeway, when the density of traffic flow reaches a high level, the traffic flow will decrease, and the density and velocity of traffic flow will oscillate, which is called the traffic wave \cite{treiber2014traffic}, i.e., the stop-and-go wave.
Traffic waves not only cause traffic jams and reduce driving comfort \cite{li2020influence,zhou2020modeling}, but also lead to increased fuel consumption \cite{wen2020mapping,shen2020does}.
To analyze this phenomenon and study the spatiotemporal distribution characteristics of the traffic flow, macroscopic traffic modeling is used, which is particularly well suited to control design since it describes the overall spatiotemporal dynamics of the traffic flow on a freeway segment. The state variables of the macroscopic model are continuous in time and space, and they can be measured and be regulated via modern traffic management system.
Lighthill, Whitham, and Richards \cite{lighthill1955kinematic, richards1956shock} proposed the LWR traffic flow model, which represents a conservation law of traffic density and describes the formation and propagation of traffic density waves on the road.
However, the oscillatory, unstable behaviors observed in the stop-and-go traffic is easier to be described by a higher-order model rather than the static LWR model.
The second-order Aw-Rascle-Zhang (ARZ) model proposed by Aw, Rascle, and Zhang \cite{aw2000resurrection,zhang2002non} adds another partial differential equation (PDE) for the velocity state, which leads to a nonlinear coupled hyperbolic PDE system.
In the literature \cite{fan2013data}, the authors investigated the extent to which the second-order ARZ traffic models could improve the prediction accuracy compared with the first-order LWR model.

Since the second-order PDE traffic flow models constitute realistic descriptions of the traffic dynamics, such as stop-and-go traffic \cite{kolb2017capacity, yu2018traffic}, boundary control designs have been recently developed for such systems \cite{belletti2015prediction, karafyllis2018feedback, kolb2017capacity, yu2019traffic, yu2018traffic, zhang2017necessary}.
In the literature \cite{yu2018traffic}, the authors developed the full-state feedback boundary control to reduce two-lane traffic congestion in a freeway segment.
In the literature \cite{yu2019traffic}, boundary feedback control of the inhomogeneous ARZ model was proposed using the infinite-dimensional backstepping method, and the boundary feedback control law of the on-ramp flow was designed to stabilize the traffic flow in the upstream of the freeway. The boundary control usually relies on static road infrastructure to regulate traffic flow, such as ramp metering and varying speed limit (VSL). With the development of adaptive cruise control (ACC) technology, in-domain control has been applied to many traffic systems \cite{Diakaki2015Overview,9018188,zheng2020smoothing}.
By controlling the time gap of ACC vehicles in the literature  \cite {Diakaki2015Overview},  the oscillation of traffic flow on the road was decreased, and the maximum capacity of the freeway could be used as much as possible to stabilize the density and velocity of the traffic flow and to alleviate traffic congestion.
A Lyapunov state feedback controller was designed in the literature  \cite{9018188}, which eliminated the traffic flow fluctuations by controlling the time gap of ACC vehicles and achieved significant optimization of the traffic flow system in terms of the fuel consumption, total travel time, and travel comfort.

The input delay of the traffic system can be derived from the transmission of information and the reaction time of a driver \cite{burger2019derivation}, which is another reason for the traffic waves.
The study in the literature  \cite{auriol2018delay} illustrates the contradictory relationship between the delay robustness and the convergence rate of the hyperbolic PDE system, and the controller is also designed to balance the convergence rate and delay robustness of the target system by using the backstepping method.
In the literature  \cite{qi2021delaycompensated}, a backstepping compensation of the in-domain controller was proposed for a same traffic model as that in the paper, which used a transport equation to represent the input delay, and the $L_2$ norm exponential stability was proven.


The Lyapunov controllers and the backstepping controller have achieved certain theoretical achievements. The Lyapunov control  stabilizes the delay-free system and the backstepping control shows excellent performance in delay compensation. However, there are still some unsolvable problems, such as Lyapunov control is not robust to time delay and the backstepping control is not robust to parameter perturbation.  Of late, reinforcement learning (RL)-based methods have been introduced to solve these problems. The RL has been used in traffic\cite{wei2019mixed, rasheed2020deep, wang2019a, yu2019reinforcement}.
In the literature  \cite{wei2019mixed}, the author used RL to compare macro and micro traffic controls and verified through simulation software that the macro traffic control could achieve better results than the micro traffic control.
In the literature  \cite{yu2019reinforcement}, a ramp metering control law based on the RL-based algorithm was proposed and compared with the boundary controller designed by the backstepping method and Lyapunov method, indicating the potential of this RL algorithm in the field of traffic control.

In this paper, we develop a RL-based controller which is trained by the the proximal policy optimization (PPO) for a traffic system compose of $2\times2$ PDEs with input delay and show the controller is robust to parameter disturbance.
Proximal policy optimization (PPO) \cite{schulman2017proximal} is chosen over other policy gradient methods, such as trust policy optimization (TRPO) \cite{schulman2017proximal} or deep deterministic policy gradient (DDPG) algorithm \cite{lillicrap2015continuous}, because PPO is more robust to the step size  and shows higher performance and lower computational complexity.
Unlike the boundary ramp metering control in the literature \cite{yu2019reinforcement}, where the PPO trains boundary inputs, i.e., one value per step, in this study, the time-gap of the ACC-equipped vehicle is used as manipulation values which are in-domain inputs. It is difficult for the PPO to train so many inputs of all the in-domain points.   Inspired by the backstepping controller but much simpler than that of the backstepping segmented controller, we take a weighted sum of three feedbacks, the feedback from the current two states and the feedback from the previous step input. The gains of the controller are trained through the interaction between the PPO and the numerical traffic system simulator.

   We design numerical simulation experiments to compare the performance of the Lyapunov control, the backstepping control and the PPO control.  For a delay-free system, the PPO control has better convergence speed and less control effort than the Lyapunov control. For a input delay system, the Lyapunov control cannot stabilize the system, while the PPO and the backstepping control show similar control performance. When the delay value is mismatched, i.e., the delay applied in control is different from the actual delay value, the performance of the PPO control is comparable to that of the backstepping control. However, only the PPO control can stabilize the system which is subject to the parameter disturbance, which shows its robustness to the parameter disturbance.

The contributions of this paper can be summarized as follows.
First, we proposed a PPO-based controller via training three normal distributed control gains to stabilize a delayed traffic flow system containing ACC-equipment vehicles. Second, although the feedback form of the PPO controller is much simpler than that of the backstepping controller, the performance of the PPO controller in  compensating for input delay is comparable to that of the backstepping delay compensator\cite{qi2021delaycompensated}. Third, the proposed PPO control is robust to parameter disturbance, in which case the backstepping method fails.

The rest of this paper is arranged as follows.
Section II introduces the traffic model.
Section III proposes the controller based on the proximal policy optimization.
Section IV includes numerical simulation experiments and comparative analysis.
Section V summarizes the paper.

\section{macroscopic traffic model}\label{model}
\begin{figure}[h]
	\centering
	\includegraphics[width=0.5\linewidth]{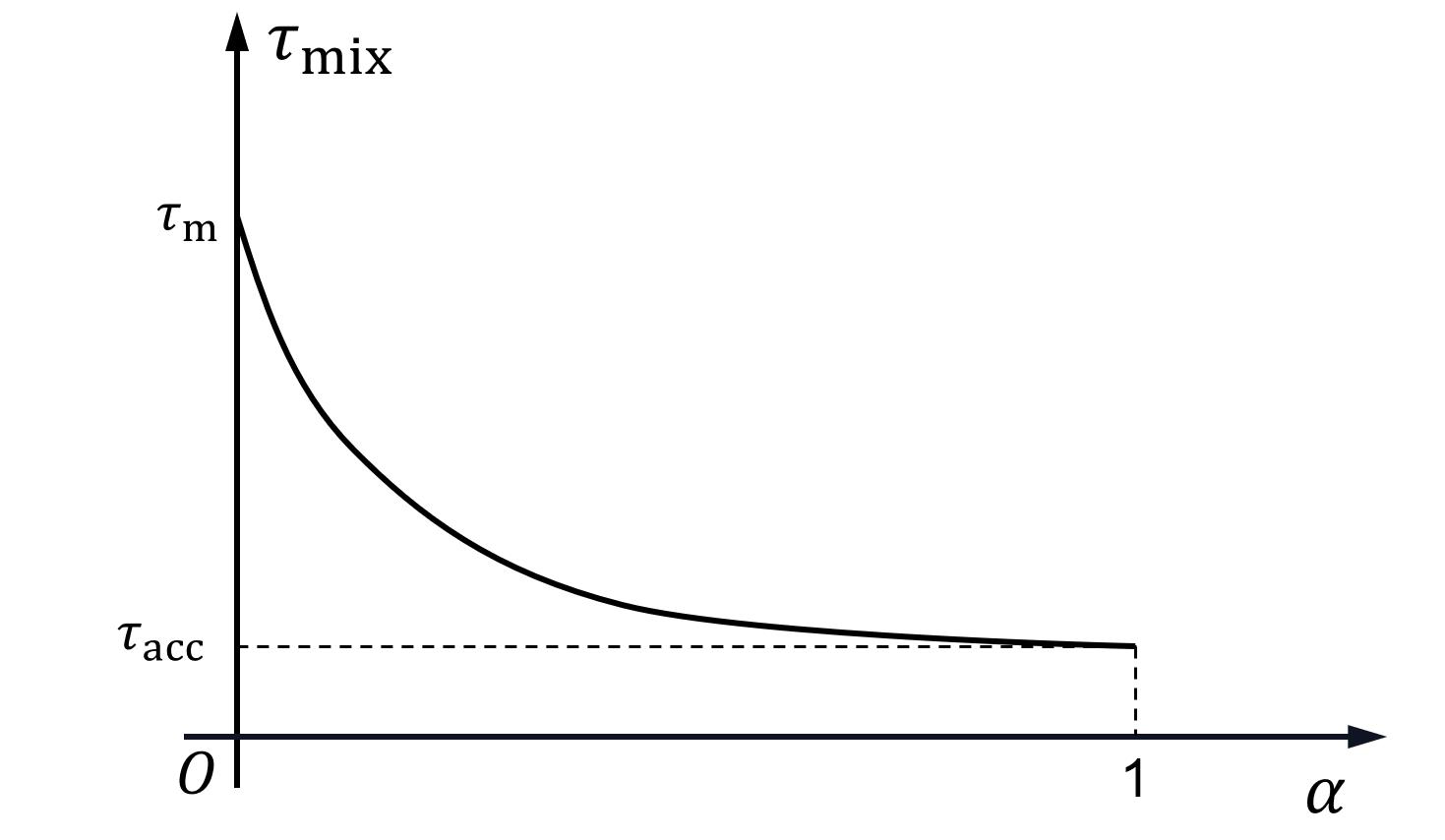}
	\caption{$\tau_{\rm mix}$ varies with $\alpha$.}
	\label{fig:tau_mix}
\end{figure}

We consider a Aw-Rascle-Zhang-type traffic model in a highway stretch $L$, proposed in the literature \cite{9018188}, described the traffic flow dynamics consisting of both adaptive cruise control-equipped (ACC-equipped) and manual vehicles. The percentage of ACC-equipped vehicles with respect to total vehicles is denoted by $\alpha \in [0,1]$. The state variables of the model are the traffic density $\rho(x, t)$ and the traffic speed $v(x, t)$, both defined in domain $(x,t)\in [0,L]\times R^+$ where $t$ is time, $x$ is the spatial variable at the concerned highway.
The traffic flow model with state variables traffic density $\rho(x,t)$ and traffic velocity $v(x, t)$ in domain $x \in[0, L], t \in\mathbb{R}^+$ is expressed as:
\begin{align}
    {\rho}_t(x,t)=&-{\rho}_x(x,t){v}(x,t)-{\rho}(x,t){v}_x(x,t),\label{traffic_1}\\
    {v}_t(x,t)=&-{\rho}(x,t)\frac{\partial V_{\rm{mix}}({\rho}(x,t),h_{\rm{acc}}(x,t-D))}{\partial {\rho}}{v}_x(x,t)\nonumber\\
    &+\frac{V_{\rm{mix}}({\rho}(x,t),{h}_{\rm{acc}}(x,t-D))-{v}(x,t)}{\tau_{\rm{mix}}}\nonumber\\
    &-{v}(x,t){v}_x(x,t),\label{traffic_2}
\end{align}
and the boundary conditions are as follow:
\begin{align}
    {v}(0,t)=&q_{\rm{in}}/{\rho}(0,t),\label{traffic_3}\\
    {v}_t(L,t)=&\frac{V_{\rm{mix}}({\rho}(L,t),{h}_{\rm{acc}}(L,t-D))-{v}(L,t)}{\tau_{\rm{mix}}},\label{traffic_4}
\end{align}
where
\begin{align}
\tau_{\rm{mix}}(\alpha) = &\frac{1}{\frac{\alpha}{\tau_{\rm{acc}}}+\frac{1-\alpha}{\tau_{\rm{m}}}}\label{tau-mix}, ~~0\leq \alpha \leq 1
\end{align}
is time constant for a mixture traffic reflecting how quickly vehicles adjust their velocity to the desired value, which depends on both
time constant $\tau_{\rm{acc}}$ of ACC vehicles and time constant $\tau_{\rm{m}}$ of
manual vehicles. $\tau_{\rm{mix}}$ also depends on $\alpha$, the percentage of ACC vehicles to total vehicles. Figure  \ref{fig:tau_mix} shows the relations of these two parameters.
Parameter $q_{\rm{in}} > 0$ is a constant external inflow.
The traffic flow observes the mass conservation law \cite{karafyllis2018feedback}, which is described by equation \eqref{traffic_1} and momentum equation derived from the velocity dynamics of ARZ model \cite{zhang2002non}, which is described by equation \eqref{traffic_2}.
To stabilize the traffic flow so as to mitigate the stop-to-go wave in the congested regime, we will design a control input actuating on the ACC-equipped vehicle by manipulation of the time-gap of ACC-equipped vehicles from their leading car, denoted by  $h_{\rm{acc}}(x,t)> 0$.
Input delays sometimes occur due to information transmission or driver's response to control instruction \cite{burger2019derivation}. Hence, we consider time delay $D>0$ in the model and design control which can compensate the input delay.

Consider the mixed traffic containing both manual and ACC-equipped vehicles, which gives the following equilibrium speed profile:
\begin{align}
    V_{\rm{mix}}({\rho},{h}_{\rm{acc}})=&\frac{1}{h_{\rm{mix}}({h}_{\rm{acc}})}\left(\frac{1}{{\rho}}-l \right ) ,\label{V-mix} ~\rho>\rho_{\rm{min}} ,
\end{align}
with mixed time gap $h_{\rm{mix}}$ is defined as follow
\begin{align}
    h_{\rm{mix}}({h}_{\rm{acc}})=&\frac{\alpha+(1-\alpha)\frac{\tau_{\rm{acc}}}{\tau_{\rm{m}}}}
    {\alpha+(1-\alpha)\frac{\tau_{\rm{acc}}}{\tau_{\rm{m}}}\frac{{h}_{\rm{acc}}}{h_{\rm{m}}}}{h}_{\rm{acc}}, \label{h-mix}
    \end{align}
where parameter  $l > 0 $ denotes the average effective vehicle length, $h_{\rm{m}}>0$ is the time gap of manual vehicles.

The effect of the penetration rate of ACC-equipped vehicles is incorporated via the mixed relaxation time \eqref{tau-mix} and the mixed time gap \eqref{h-mix}, with $h_{\rm{min}}<\min\{h_{\rm{m}},h_{\rm{acc}}\}\le h_{\rm{mix}}\le \max\{h_{\rm{m}},h_{\rm{acc}}\}<h_{\rm{max}}$, where $h_{\rm{min}}$ and $h_{\rm{max}}$ denote the minimal and the maximal value of $h$, respectively, for all $\alpha\in[0,1]$.

\begin{figure}[h]
    \centering
    \includegraphics[width=0.45\linewidth]{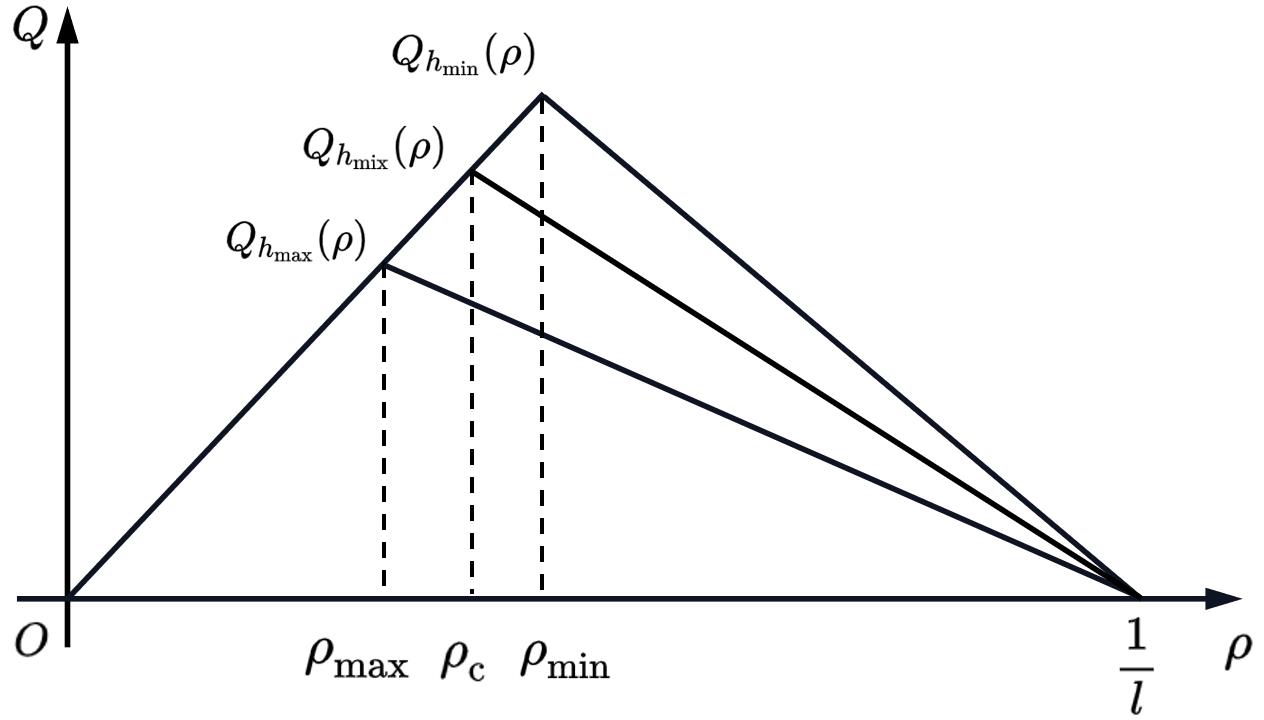}
    \caption{Fundamental diagram.}
    \label{fig:fundamental}
\end{figure}

Figure \ref{fig:fundamental} shows the relationship of mixed traffic flow $Q$ and the traffic density $\rho$, where $\rho_c>0$ is the critical value of the traffic density, that is, when the density is greater than it, the traffic becomes congested.

The traffic flow $Q_{h_{\rm{mix}}}$ is defined as
\begin{align}\label{Q-mix}
    Q_{h_{\rm{mix}}}(\rho)=
    \begin{cases}
        &v_{\rm{f}}\rho,~~ 0\le \rho \le \rho_{\rm{c}},\\
        &\frac{1-l\rho}{h_{\rm{mix}}},~~ \rho_{\rm{c}} < \rho \leq \frac{1}{l}, \\
    \end{cases}
\end{align}
where $\rho_{\rm{c}} = 1/(l+h_{\rm{mix}}v_{\rm{f}})$ and $v_{\rm{f}}$ is free-flow speed. Since we know $V_{\rm{mix}}(\rho,h_{\rm{acc}})=Q_{h_{\rm{mix}}}/\rho$, $0<V_{\rm{mix}}({\rho},{h}_{\rm{acc}})<v_{\rm{f}}$ for all $\alpha \in [0,1]$ and $\rho_{\rm{min}}< \rho < (1/l)$ is guaranteed by $\max\{h_{\rm{acc}},h_{\rm{m}}\}\le h_{\rm{max}}$ and $\min\{h_{\rm{acc}},h_{\rm{m}}\}\ge h_{\rm{min}}$. Hence, equation \eqref{V-mix} defines a reasonable fundamental diagram for mixed traffic in congested conditions.
Using an analysis method similar to the one employed in the literature
\cite{zhang2002non}, one can find that system \eqref{traffic_1}-\eqref{traffic_4} is anisotropic.

Given an inflow $ q_{\rm{in}} $ and a constant time gap input $ \bar{h}_{\rm{acc}} $, we find the equilibrium of system \eqref{traffic_1}-\eqref{traffic_4} as follows, which is same as that in the literature \cite{9018188},
\begin{align}\label{steady-state}
        \bar{v}=\frac{l}{(1/q_{\rm{in}})-\bar{h}_{\rm{mix}}},~~~\bar{\rho}=\frac{1}{l+\bar h_{\rm{mix}}\bar v},
\end{align}
with mixed time gap
\begin{align}
    \bar{h} _{\rm{mix}}=\frac{\alpha+(1-\alpha)\frac{\tau_{\rm{acc}}}{\tau_{\rm{m}}}}    {\alpha+(1-\alpha)\frac{\tau_{\rm{acc}}}{\tau_{\rm{m}}}\frac{\bar{h}_{\rm{acc}}}{h_{\rm{m}}}}\bar{h}_{\rm{acc}}.\label{bar-h_acc}
\end{align}

The open-loop system is unstable to the equilibrium \eqref{steady-state}.  The Lyapunov feedback control proposed in the literature  \cite{9018188} can stabilize the linearized system without delay, and the backstepping control designed in the literature  \cite{qi2021delaycompensated} can stabilize the linearized system with input delay.
In the next section, we will design a RL-based controller under the states feedback and historical input feedback framework, where its control policy is trained by the PPO.

\section{PPO  control design}\label{PPO}

PPO is a model-free, policy-gradient method, whose hyperparameters are robust to multiple tasks. Also, PPO algorithm exhibits high performance and low computational complexity \cite{schulman2017proximal} compared to other policy gradient methods. The explicit knowledge of the traffic model is not required for PPO-based control approaches. In the paper, we simulate traffic dynamics by discretizing the nonlinear model by the finite difference method and use this numerical simulator as the environment with which the PPO interacts. The solution $\rho(x,t)$ and $v(x,t)$ to \eqref{traffic_1}-\eqref{traffic_4} is discretized in domain $[0,L]\times [0,T]$. The discretization space step is denoted by $\Delta x=L/M$ and time step is denoted by $\Delta t=T/N$, which are chosen such that the Courant-Friedrichs-Lewy condition is met, namely $\frac{\Delta x}{\Delta t} \geq \max|\lambda_{1,2}|$, where $\lambda_{1,2}$ are the characteristic speed of the model \eqref{traffic_1}-\eqref{traffic_4}. Hence, the discretized states and input at time $t$ can be written as:
\begin{align}
	\boldsymbol{\rho}^{t}=&[\rho(x_0,t),\rho(x_1, t),...,\rho(x_m,t)],\\
	\boldsymbol{v}^{t}=&[v(x_0,t),v(x_1, t),...,v(x_m,t)],\\
	\boldsymbol h_{\rm{acc}}^{t-1}=&[h_{\rm{acc}}(x_0,t-1),h_{\rm{acc}}(x_1,t-1),...,\nonumber\\
	&~h_{\rm{acc}}(x_m,t-1)],
\end{align}
where $x_i=\Delta x\times i$ for $i=0,...,M$ and $t=\Delta t\times j$ for $i=0,...,N$.

To avoid too many output of the policy network which may cause the training of the weights to diverge, we employ the PPO to learn the control gains of the feedback, rather than all the inputs distributed in domain. The control form is inspired by the backstepping delay compensator in the literature  \cite{qi2021delaycompensated}, which consists three components:
\begin{align}
	&\boldsymbol h_{\rm{acc}}^{t}=-\eta_1^t(\boldsymbol h_{\rm{acc}}^{t-1}-\boldsymbol{\bar{h}}_{\rm{acc}})-\eta_2^t(\boldsymbol v^t-\boldsymbol{\bar{v}})+\eta_3^t(\boldsymbol\rho^t-\boldsymbol{\bar{\rho}}).\label{h_acc}
\end{align}
The control includes the feedback of the current velocity and density states and the previous one step input, which is much simpler in form than that of the backstepping control \cite{qi2021delaycompensated}.

Consider an infinite-horizon discounted Markov decision process, defined by the tuple ($\mathcal{S}$, $\mathcal{A}$, $\mathcal{P}$, $\mathcal{R}$) representing the dynamics of the traffic flow.

$\mathcal{S}$: Space of states of the environment, which are the input of the policy network. Let $s^t\in \mathcal{S}$ be a state at time $t$ defined as
\begin{align}\label{states}
    \boldsymbol{s}^{t}=&[\boldsymbol{\rho}^{t},\boldsymbol{v}^{t},\boldsymbol h_{\rm{acc}}^{t-1}].
\end{align}

$\mathcal{A}$: Space of actions. At each time $t$, the control policy choose an action $\boldsymbol{a}^t\in\mathcal{A}$, which is the input of the environment. Define $\boldsymbol{a}^{n}$ as follows:
\begin{align}\label{action}
    \boldsymbol{a}^{t}=&[\eta_1^{t},\eta_2^{t},\eta_3^{t}].
\end{align}

Policy:
Let $\pi(\boldsymbol{a}|\boldsymbol{s})$ denote a policy function $\pi:\mathcal{S}\times\mathcal{A}\to[0,1]$, which represents the probability of taking an action $\boldsymbol{a}\in\mathcal{A}$ given state $\boldsymbol{s}$.  In the control learning, we suppose the policy functions are the three normal distributions of three gains, respectively:
\begin{align}
    \eta_1^{t}\sim \mathcal{N}(\mu_1^{t},\sigma_1^{t}),\\
    \eta_2^{t}\sim \mathcal{N}(\mu_2^{t},\sigma_2^{t}),\\
    \eta_3^{t}\sim \mathcal{N}(\mu_3^{t},\sigma_2^{t}),
\end{align}
where $\mu_i^t$ and $\sigma_i^t$ are the mean and the variance for $\eta_i^t$, respectively. The policy-based method applied, we defined a policy network  $g_{\rm{DNN}}$ to approximate the policy functions:
\begin{align}
	[\boldsymbol{\mu}^t,\boldsymbol{\sigma}^t] = g_{\rm{DNN}}(a^t|\boldsymbol{s}^{t};\theta),\label{g_dnn}
\end{align}
where
\begin{align}
	&\boldsymbol{\mu}^n=[\mu_1^{n},\mu_2^{n},\mu_3^{n}],~~
	\boldsymbol{\sigma}^n=[\sigma_1^{n},\sigma_2^{n},\sigma_3^{n}].
\end{align}
The outputs of the network are the means and variances of the three gains. In practice, the action is drawn from sampling and then the control input is calculated according to \eqref{h_acc}.

$\mathcal{P}(\boldsymbol{s}^{t+1}|\boldsymbol{s}^{t},\boldsymbol{a}^{t})$:
The state-transition probability. It defines the probability that the state $\boldsymbol{s}^{t}$ transforms to $\boldsymbol{s}^{t+1}$ under action $\boldsymbol{a}^{t}$. The randomness of the state transition is from the environment. In the paper, the environment is a simulator of the traffic model which can be written as:
\begin{align}
	\boldsymbol{s}^{t+1} = f_{\rm{traffic}}(\boldsymbol{s}^{t},\boldsymbol{h}_{\rm{acc}}^{t-d}),
\end{align}
where $d=D/\Delta t$ representing the input delay. If $D>0$, the control input will not work until $d$ of time $\Delta t$. The function $f_{\rm{traffic}}$ represents the numerical simulated dynamics for the temporal evolution of the traffic model \eqref{traffic_1}-\eqref{traffic_4}. Although the traffic dynamics is deterministic, we still generally express the state transition process by a Markov decision process probability
\begin{align}
	\boldsymbol{s}^{t+1} \sim\mathcal{P}(\boldsymbol{s}^{t+1}|\boldsymbol{s}^{t},\boldsymbol{a}^{t}).
\end{align}
The relation between $\boldsymbol{h}_{\rm{acc}}^{t-d}$ and $\boldsymbol{a}^{t}$ can be found in \eqref{h_acc}, \eqref{action} and \eqref{g_dnn}.

$\mathcal{R}(\boldsymbol{s}^{t})$: The total discounted return from time $t$ to $t+j$ can be expressed as:
\begin{align}\label{Reward}
    \mathcal{R}(\boldsymbol{s}^{t})=&r^t+\gamma r^{t+1}
    +...+\gamma^{j}r^{t+j}=\sum_{i=0}^{j}\gamma^{i} r^{t+i},
\end{align}
where $\gamma\in(0,1]$ is the discount factor that encodes the importance of future rewards, and $r^t$ is the reward depending on $\boldsymbol{s}^{t}$ as follows:
\begin{align}
    r(\boldsymbol{s}^{t})=&-\sum_{i=0}^{N}\left[\frac{\rho(x_i,t)-\bar \rho}{\bar\rho}\right]^2+\left[\frac{v(x_i,t)-\bar v}{\bar v}\right]^2.\label{reward}
\end{align}

Thus, given $\boldsymbol{s}^{t}$, the return $ \mathcal{R}(\boldsymbol{s}^{t})$
depends on   actions $\boldsymbol{a}^{t}$, $\boldsymbol{a}^{t+1}$, $\boldsymbol{a}^{t+2}$, $...$ and states
 $\boldsymbol{s}^{t+1}$, $\boldsymbol{s}^{t+2}$, $\boldsymbol{s}^{t},...$.

Value functions:  First we define the action-value function for policy $\pi$ given the current state $\boldsymbol{s}^{t}$ and action $\boldsymbol{a}^{t}$
\begin{align}
	Q_\pi(\boldsymbol{s}^{t},\boldsymbol{a}^{t})=&\mathbb E_\pi[\mathcal{R}(\boldsymbol{s}^{t})|\boldsymbol{s}^{t},\boldsymbol{a}^{t}],\label{Q_pi}
\end{align}
 which represents the expected total discounted return on actions $\boldsymbol{a}^{t}$, $\boldsymbol{a}^{t+1}$, $\boldsymbol{a}^{t+2}$, $...$ and states
 $\boldsymbol{s}^{t+1}$, $\boldsymbol{s}^{t+2}$, $\boldsymbol{s}^{t},...$. The probability of actions obeys the policy function and the randomness of the states come from the state-transition probability. Accordingly, we define the optimal action-value function as
 \begin{align}
	Q^*(\boldsymbol{s}^{t},\boldsymbol{a}^{t})=Q_\pi(\boldsymbol{s}^{t},\boldsymbol{a}^{t}).\label{Q_star}
\end{align}
Then, the state-value function is
\begin{align}
V_{\pi}(\boldsymbol{s}^{t})=\mathbb E_{\mathcal{A}}[Q_\pi(\boldsymbol{s}^{t},\mathcal{A}]
\end{align}
which is the expected value of action-value function on all actions in $\mathcal{A}$.
The advantage function, reflecting the how advantageous the action $\boldsymbol{a}^{t}$ compared to the expectation of all actions in $\mathcal{A}$, is defined as:
\begin{align}
	A_\pi(\boldsymbol{s}^{t},\boldsymbol{a}^{t})=&Q_\pi(\boldsymbol{s}^{t},\boldsymbol{a}^{t})-V_\pi(\boldsymbol{s}^{t}),\label{A_pi}
\end{align}

Actor-Critic Method:
The Actor-Critic method combines the policy learning and value learning, where two neural networks are used to approximate the policy function and the state-value function, receptively.

First, we consider the policy network (actor network), denoted by  $\pi_\theta(\boldsymbol{a}^{t}|\boldsymbol{s}^{t})$, where $\theta$ is the parameters of the neural network which is updated by using the PPO method. The PPO allows for continuous states and action space, which trains a continuous-valued stochastic control policy.  There are two policy networks running in the learning process simultaneously, one is training network, denoted by $\pi_{\rm{\theta}}$; the other is interaction network denoted by $\pi_{{\rm{\theta}}_{\rm{old}}}$. The interaction network interact with the environment so as to collect the data of a group of actions and the corresponding states. These data is necessary for the training network to update its parameters. After collecting enough data, network $\pi_{{\rm{\theta}}_{\rm{old}}}$ updates its parameters according to $\pi_\theta(\boldsymbol{a}^{t}|\boldsymbol{s}^{t})$.

The policy gradient is estimated via optimizing:
\begin{align}
	\theta' = \arg \max_\theta J(\theta),\label{theta}
\end{align}
where the objective function is
\begin{align}
    J(\theta)=
    &\mathbb{E}_\pi[\min\{g^{t}(\theta),
    \mathrm{clip}(g^{t}(\theta),1-\epsilon,1+\epsilon)\}\nonumber\\
    &\times A_\pi(\boldsymbol{s}^{t},\boldsymbol{a}^{t})],\label{J}
\end{align}
with
\begin{align}
    g^{t}(\theta)=\frac{\pi_{\theta}(\boldsymbol{s}^{t}|\boldsymbol{a}^{t})}{\pi_{\theta_{\rm{old}}}(\boldsymbol{s}^{t}|\boldsymbol{a}^{t})}.\label{gt}
\end{align}
The first term of  \eqref{J} inside the min is $g^{n}$. The second term, $\mathrm{clip}(g^{t}(\theta),1-\epsilon,1+\epsilon)\}\times A_\pi(\boldsymbol{s}^{t},\boldsymbol{a}^{t})$  clips the probability ratio, which removes the incentive for moving $g^t$ outside of the interval $[1-\epsilon,1+\epsilon]$ with the hyperparameter $\epsilon$. The clipped objective \eqref{J} improves the convergence of the algorithm.
The importance sampling is applied to increase sample efficiency, in which the expectation of $\pi_{\theta}$ is computed by sampling the old policy $\pi_{\theta_{\rm{old}}}$.

To train the policy network, the value function $V$ is required to supervises the update of the policy parameters. As the value function is unknown, we use a critic network to approximate to the state-value function as follows:
\begin{align}
	V_{\pi}(\boldsymbol{s}^{t}) = f_{DNN}(\boldsymbol{s}^{t},\omega),\label{V_pi}
\end{align}
where $\omega$ is the trainable parameter of the network, which is trained by minimizing the loss function defined as:
\begin{align}
	\mathcal L(\omega) = \frac{1}{\epsilon}\sum_{i=t-\epsilon}^t(\mathcal{R}(\boldsymbol{s}^{i})-V_{\pi}(\boldsymbol{s}^{i}))^2.
\end{align}
A deep neural network (DNN) $f_{DNN}(\boldsymbol{s}^{t},\omega)$ is used as the critic.

\begin{figure}[t]
	\centering
	\includegraphics[width=0.5\linewidth]{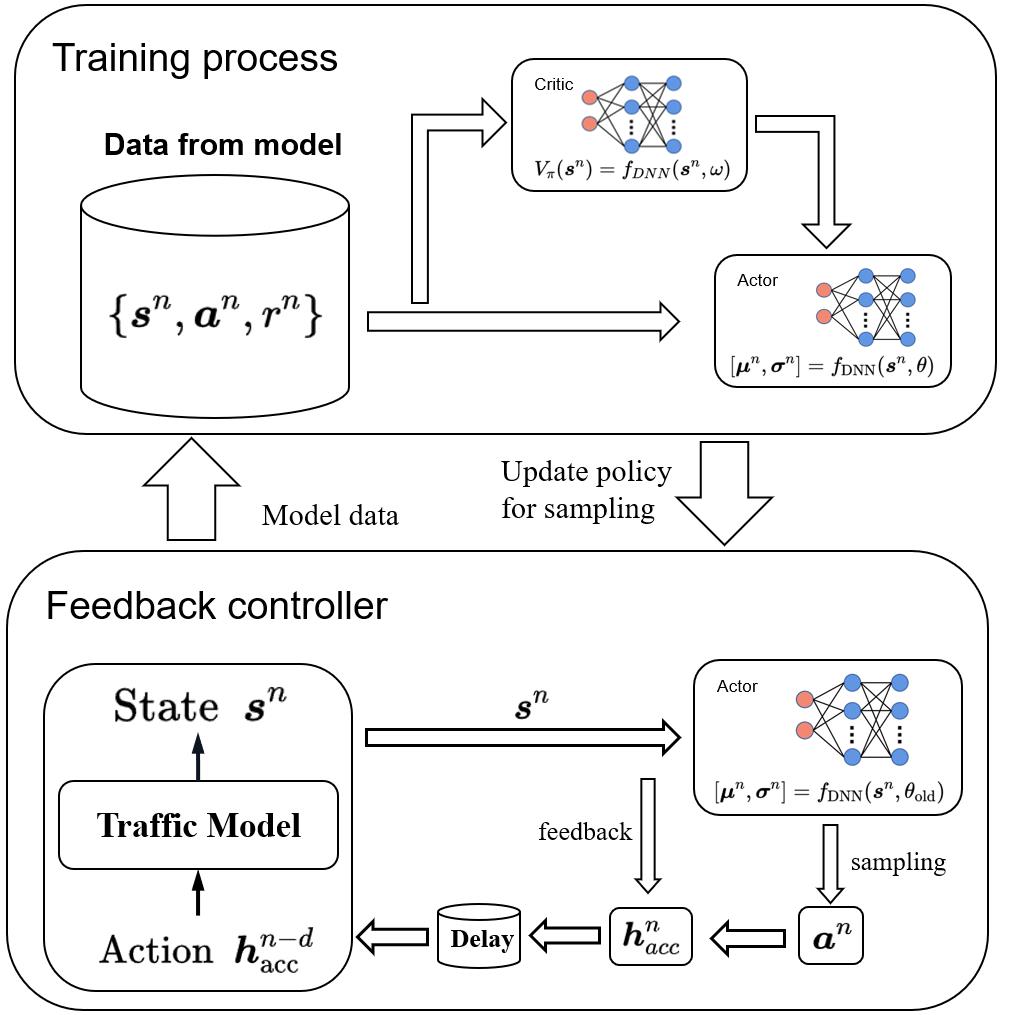}
	\caption{The proximal policy optimization in-domain control scheme.}
	\label{fig:diagram}
\end{figure}

Figure \ref{fig:diagram} illustrates the learning process of the PPO-based control.
At each time $t=0,1,...,M$, input state $\boldsymbol{s}^{t}$ into the policy network $\pi_{\theta_{\rm{old}}}$, which results in the distributions for the three control gains. Sampling the control gains from the distributions, namely action $\boldsymbol{a}^{t}$, the control $\boldsymbol{h}_{\rm{acc}}^{t}$ is calculated from equation \eqref{h_acc}.
The delayed input $\boldsymbol{h}_{\rm{acc}}^{t-d}$ is feed to the environment, namely, the discretized PDE model, which results in the states $\boldsymbol{s}^{t+1}$.
Iterating the process $E=\Delta T/\Delta t$ times, all the $M$ states and actions are collected and then input into the policy network  $\pi_{\theta}$ for training its parameters.
After update the parameters of the network $\pi_{\theta}$ using the PPO gradient estimation, the new parameter of network $\pi_{\theta}$ is transmitted to network $\pi_{\theta_{\rm{old}}}$.
The implementation of the PPO based controller is summarized in Algorithm 1.

\begin{algorithm}[htb]
\caption{Proximal Policy Optimization Procedure}
\label{alg:Framwork}
\begin{algorithmic}
\State ~1: Initialize parameters for states, actors, networks and $\epsilon= T/\Delta t$.
\State ~2: ~~\textbf{for} $n=0,1,2,3,...$ \textbf{do}
\State ~3: ~~~~\textbf{for} $t= 0$ to $ \epsilon $ \textbf{do}
\State ~4: ~~~~Set $t'=n\times\epsilon + t$.
\State ~5: ~~~~Update  $\boldsymbol{s}^{t'}$ to policy $\pi_{\theta_{\rm{old}}}$  and sampling to get $\boldsymbol{a}^{t'}$.
\State ~6: ~~~~Calculate $\boldsymbol{h}_{\rm{acc}}^{t'}$ from \eqref{h_acc} and push $\boldsymbol{h}_{\rm{acc}}^{t'}$ to the delay buffer.
\State ~7: ~~~~Get $\boldsymbol{h}_{\rm{acc}}^{t'-d}$ from the delay buffer and get state data from traffic system $\boldsymbol{s}^{t'+1} =
f_{\rm{traffic}}(\boldsymbol{s}^{t'},\boldsymbol{h}_{\rm{acc}}^{t'-d})$.
\State ~8: ~~~~\textbf{end for}
\State ~9: ~~~~Collect set of trajectories driven by policy $\pi_{\theta_{\rm{old}}}$.
\State ~10: ~~Compute total discounted reward $\mathcal R$.
\State ~11: ~~Compute advantage estimates, $A_{\pi}$ from critic $V_{\pi}$.
\State ~12: ~~Update the actor, $\theta$ by Adam optimizer.
\State ~13: ~~Update the critic, $\omega$ by Adam optimizer.
\State ~14: ~~$\theta_{\rm{old}}\leftarrow\theta$
\State ~15: ~~\textbf{end for}
\end{algorithmic}
\end{algorithm}

\begin{figure*}[t]
	\centering
	\subfigure[]{
		\begin{minipage}{0.35\linewidth}
			\includegraphics[width=1\linewidth]{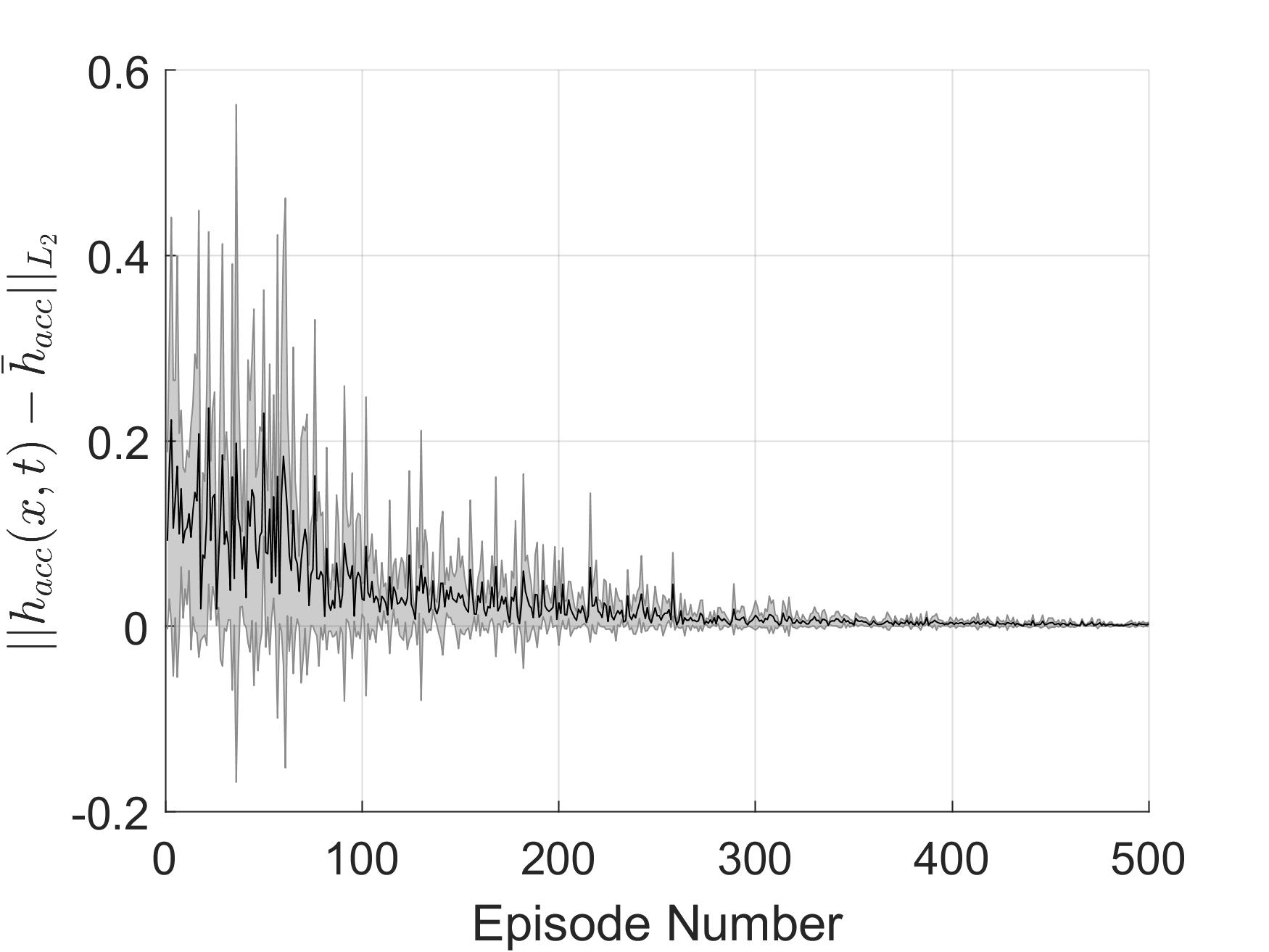}
		\end{minipage}%
	}
	\subfigure[]{
		\begin{minipage}{0.35\linewidth}
			\includegraphics[width=1\linewidth]{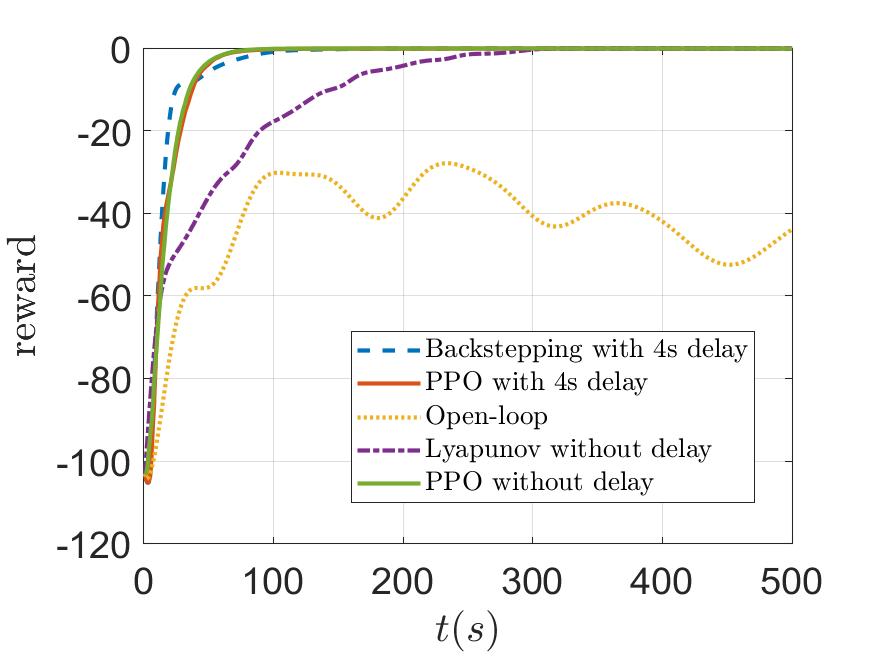}
		\end{minipage}%
	}
	\vspace{-0.3cm}
	\caption{(a) The learning curve of the PPO  controller with a 4-s input delay. (b) The reward evolution for the open-loop control system, closed-loop system with the PPO  controller, backstepping controller, and Lyapunov controller.}
	\label{fig:process}
\end{figure*}

\begin{figure*}[b]
	\centering
	\subfigure[]{
		\begin{minipage}[ht]{0.25\linewidth}
			\includegraphics[width=1\linewidth]{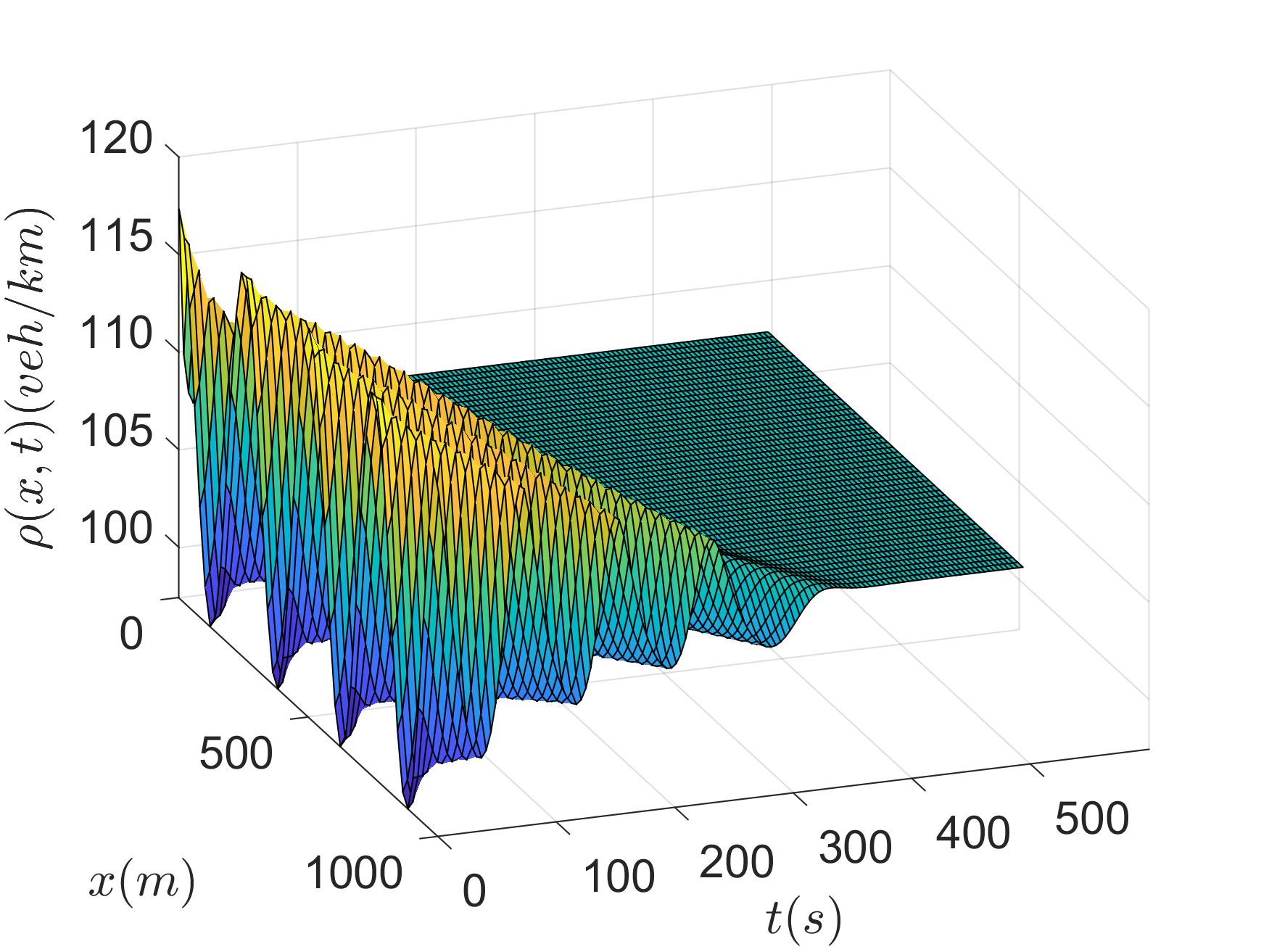}
		\end{minipage}
	}
	\subfigure[]{
		\begin{minipage}[ht]{0.25\linewidth}
			\includegraphics[width=1\linewidth]{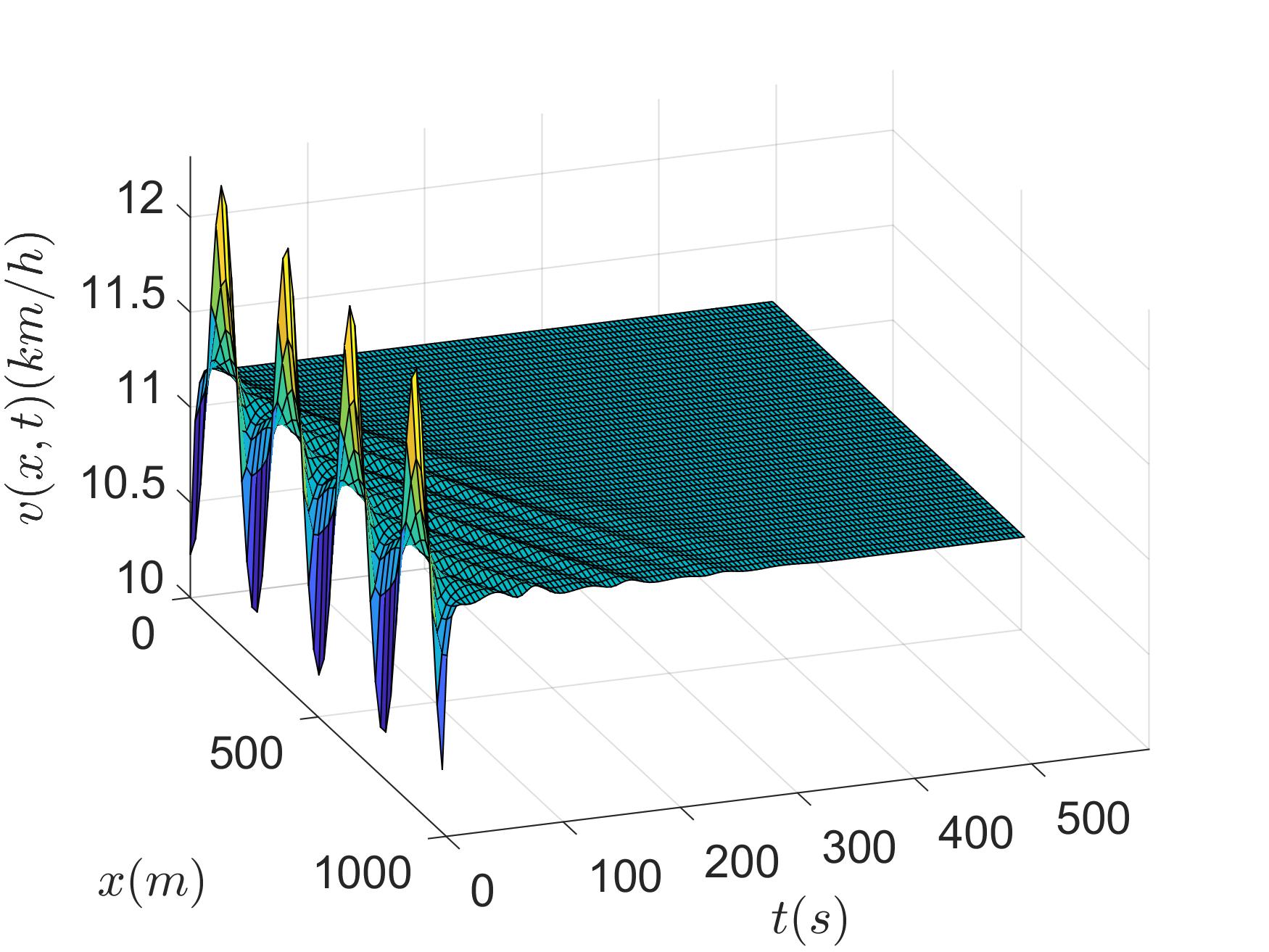}
		\end{minipage}
	}
	\subfigure[]{
		\begin{minipage}[ht]{0.25\linewidth}
			\includegraphics[width=1\linewidth]{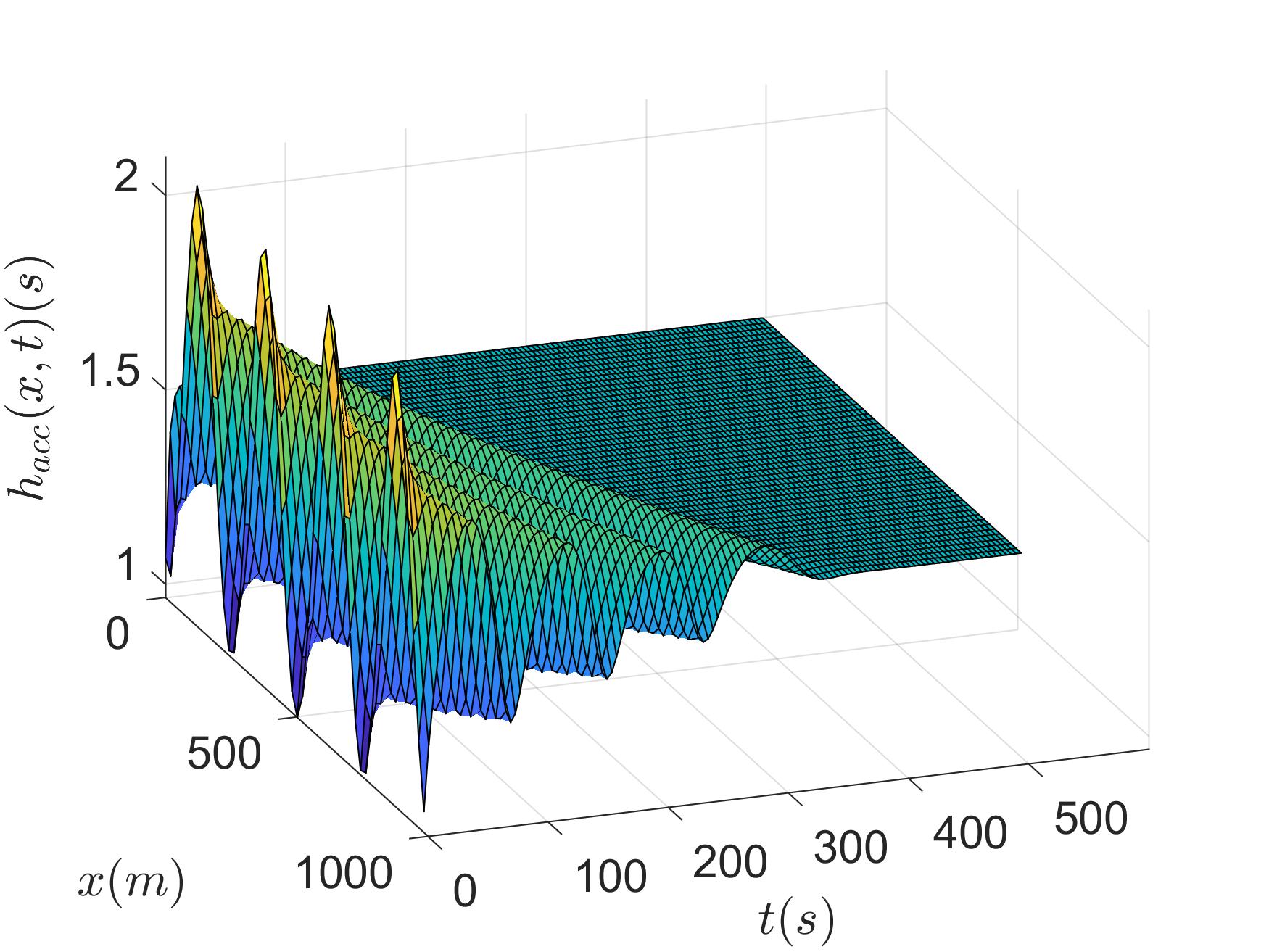}
		\end{minipage}%
	}
	\vspace{-0.3cm}
	
	\subfigure[]{
		\begin{minipage}[ht]{0.25\linewidth}
			\includegraphics[width=1\textwidth]{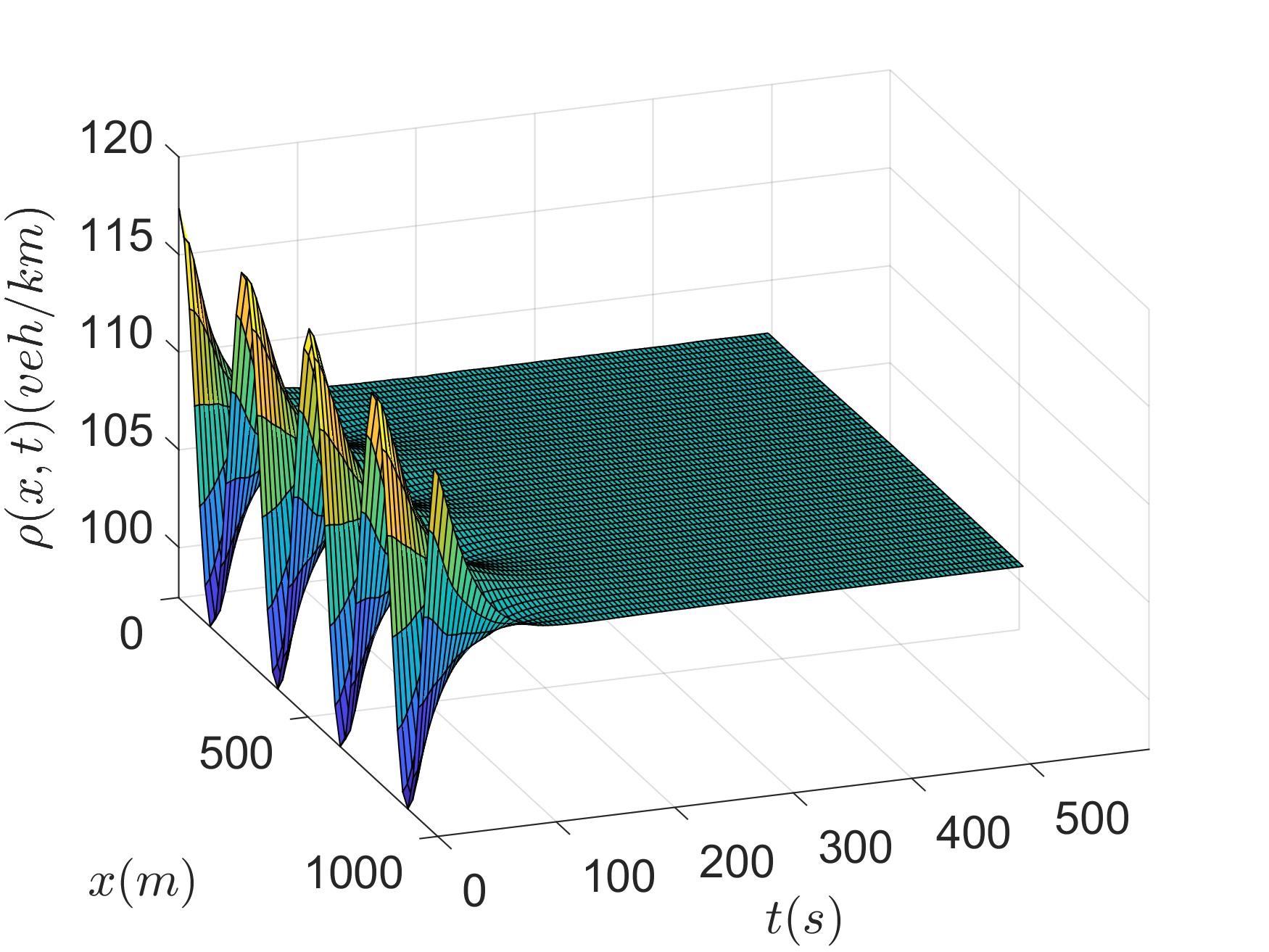}
		\end{minipage}
	}
	\subfigure[]{
		\begin{minipage}[ht]{0.25\linewidth}
			\includegraphics[width=1\linewidth]{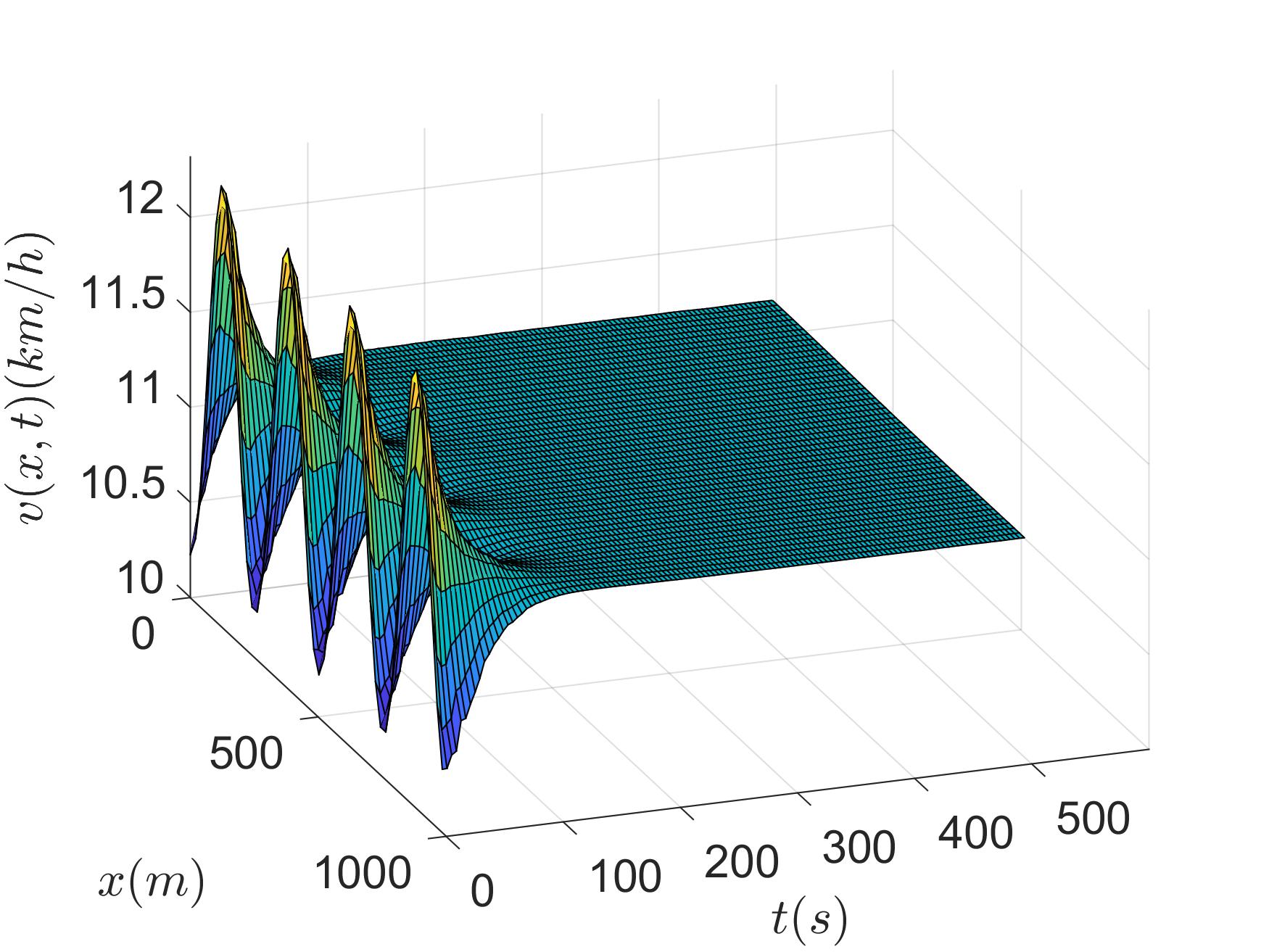}
		\end{minipage}
	}
	\subfigure[]{
		\begin{minipage}[ht]{0.25\linewidth}
			\includegraphics[width=1\linewidth]{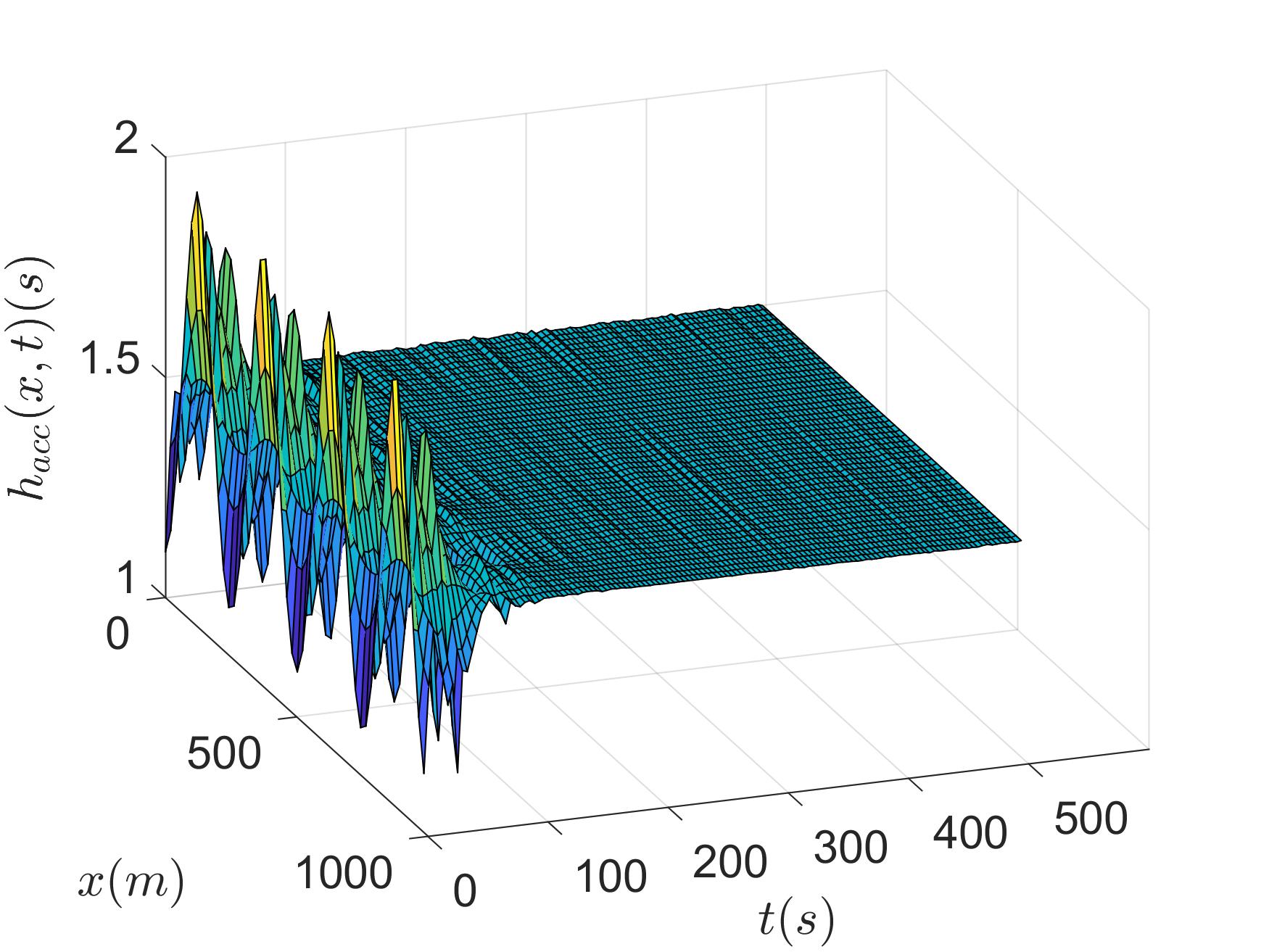}
		\end{minipage}%
	}
	\vspace{-0.3cm}
	\caption{
		(a)–(c) The states and control input of a delay free system under the Lyapunov controller \cite{9018188}; (d)–(f) The state and control input of a delay free system under the PPO control.
	}
	\label{fig:delay0s}
\end{figure*}

\begin{figure*}[t]
	\centering
	\subfigure[]{
		\begin{minipage}[ht]{0.25\linewidth}
			\includegraphics[width=1\linewidth]{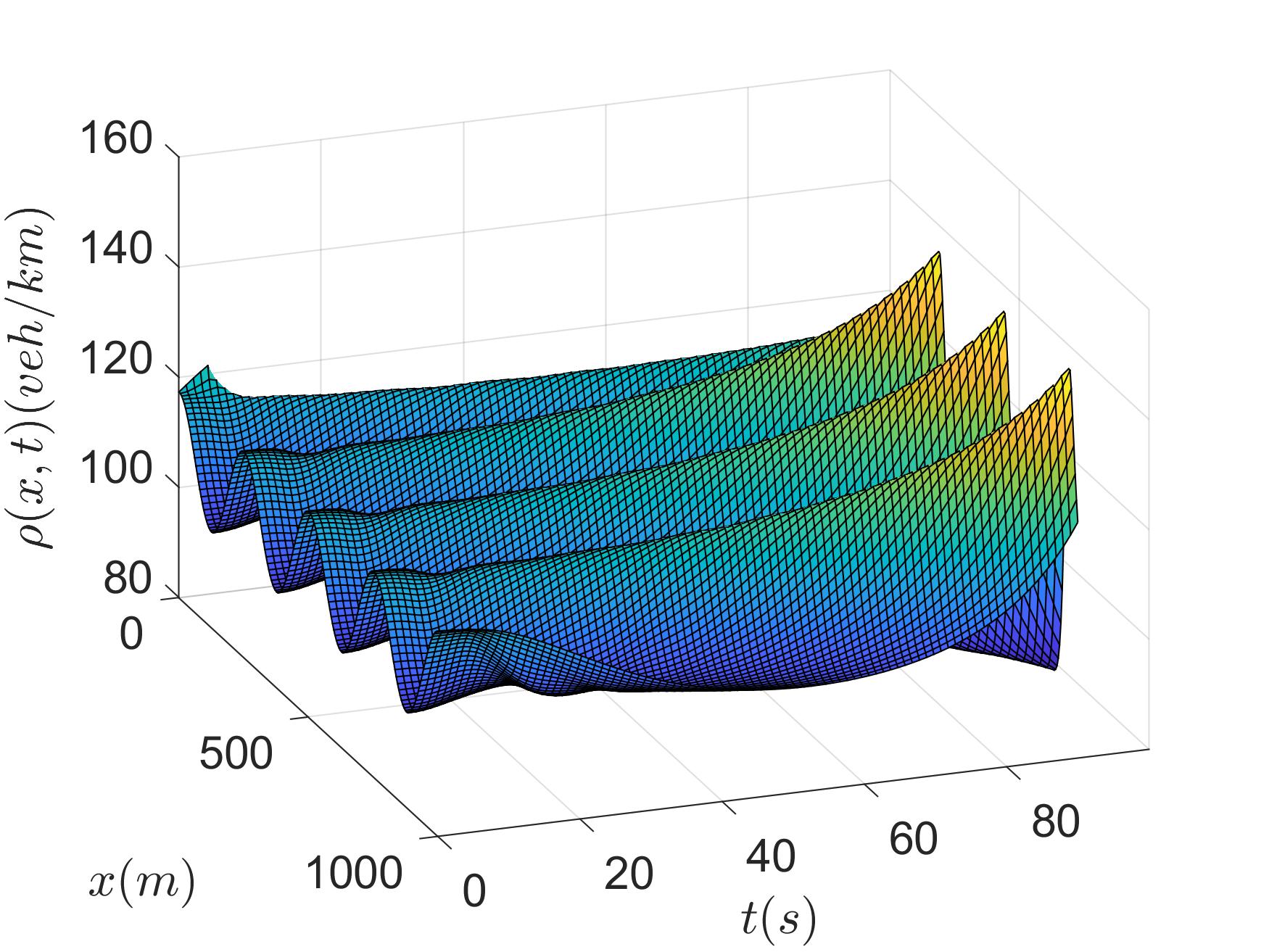}
		\end{minipage}
	}
	\subfigure[]{
		\begin{minipage}[ht]{0.25\linewidth}
			\includegraphics[width=1\linewidth]{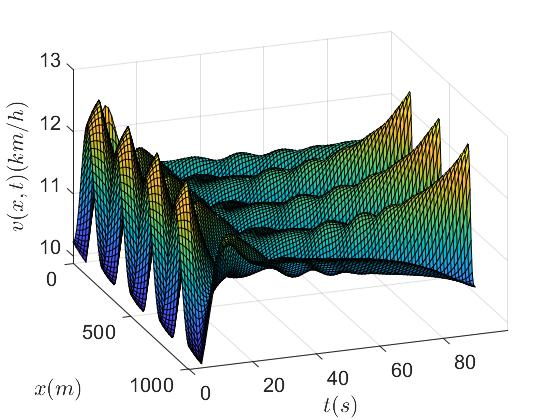}
		\end{minipage}
	}
	\subfigure[]{
		\begin{minipage}[ht]{0.25\linewidth}
			\includegraphics[width=1\linewidth]{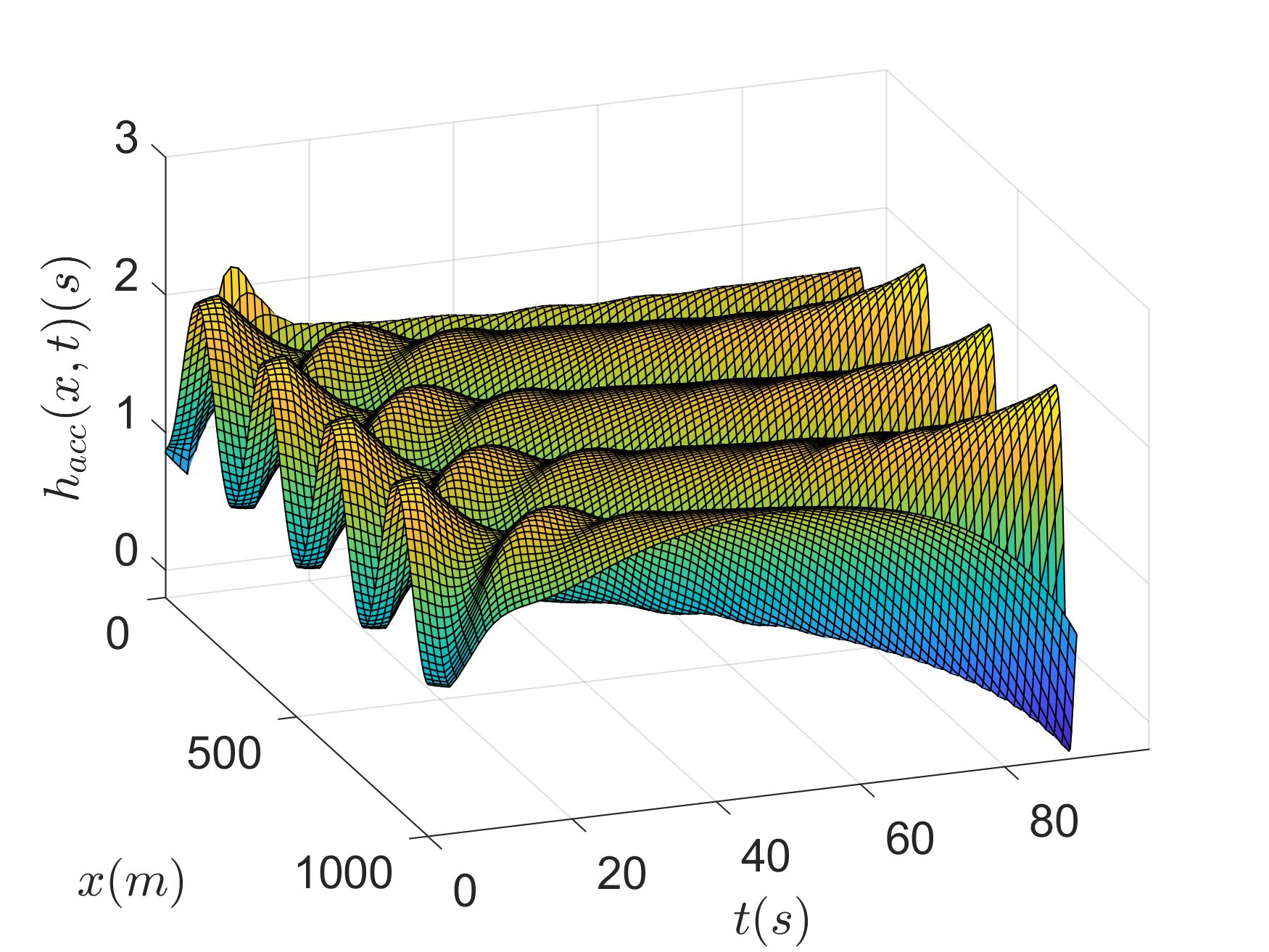}
		\end{minipage}%
	}
	\vspace{-0.3cm}
	
	\subfigure[]{
		\begin{minipage}[ht]{0.25\linewidth}
			\includegraphics[width=1\linewidth]{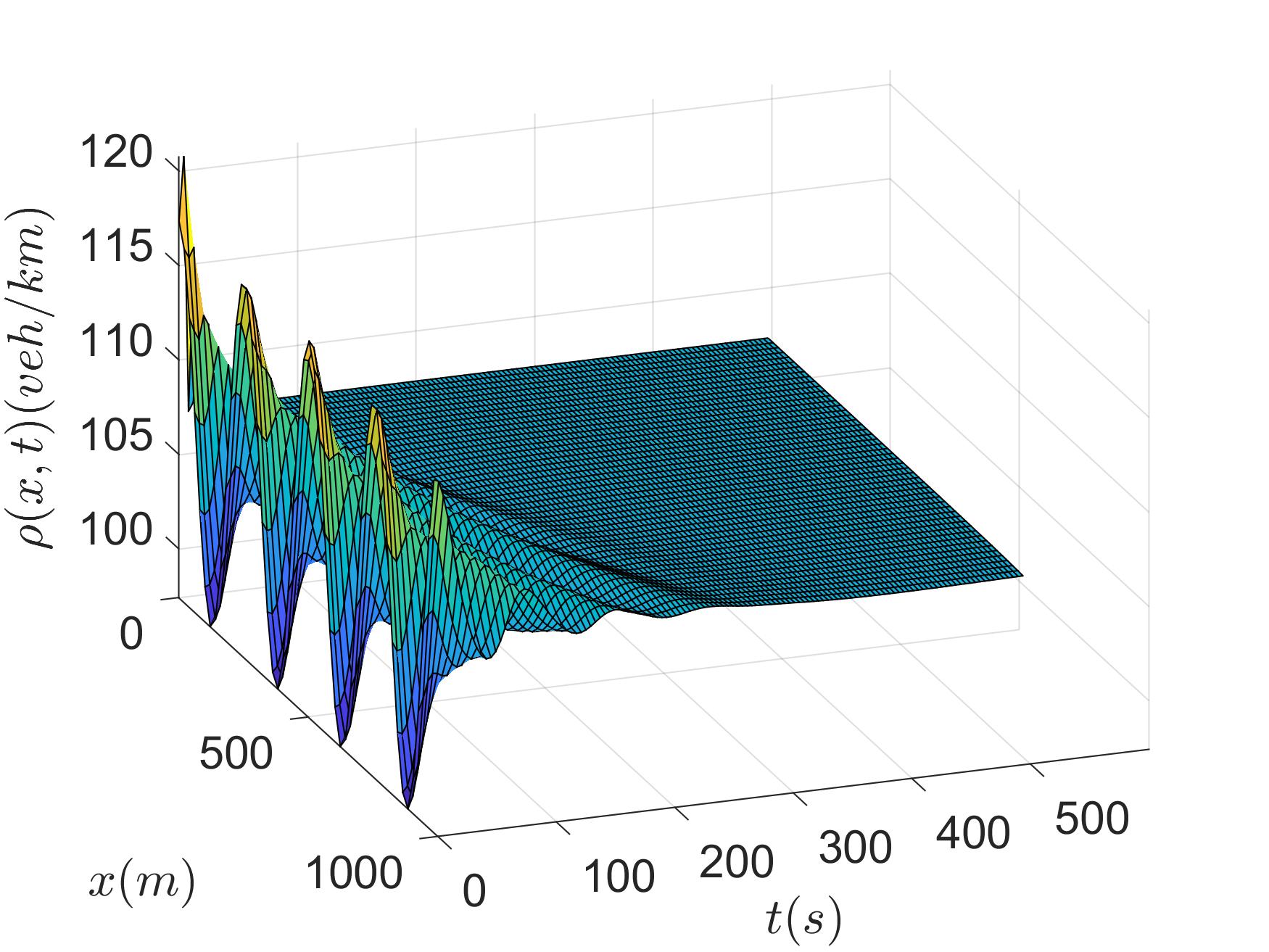}
		\end{minipage}
	}
	\subfigure[]{
		\begin{minipage}[ht]{0.25\linewidth}
			\includegraphics[width=1\linewidth]{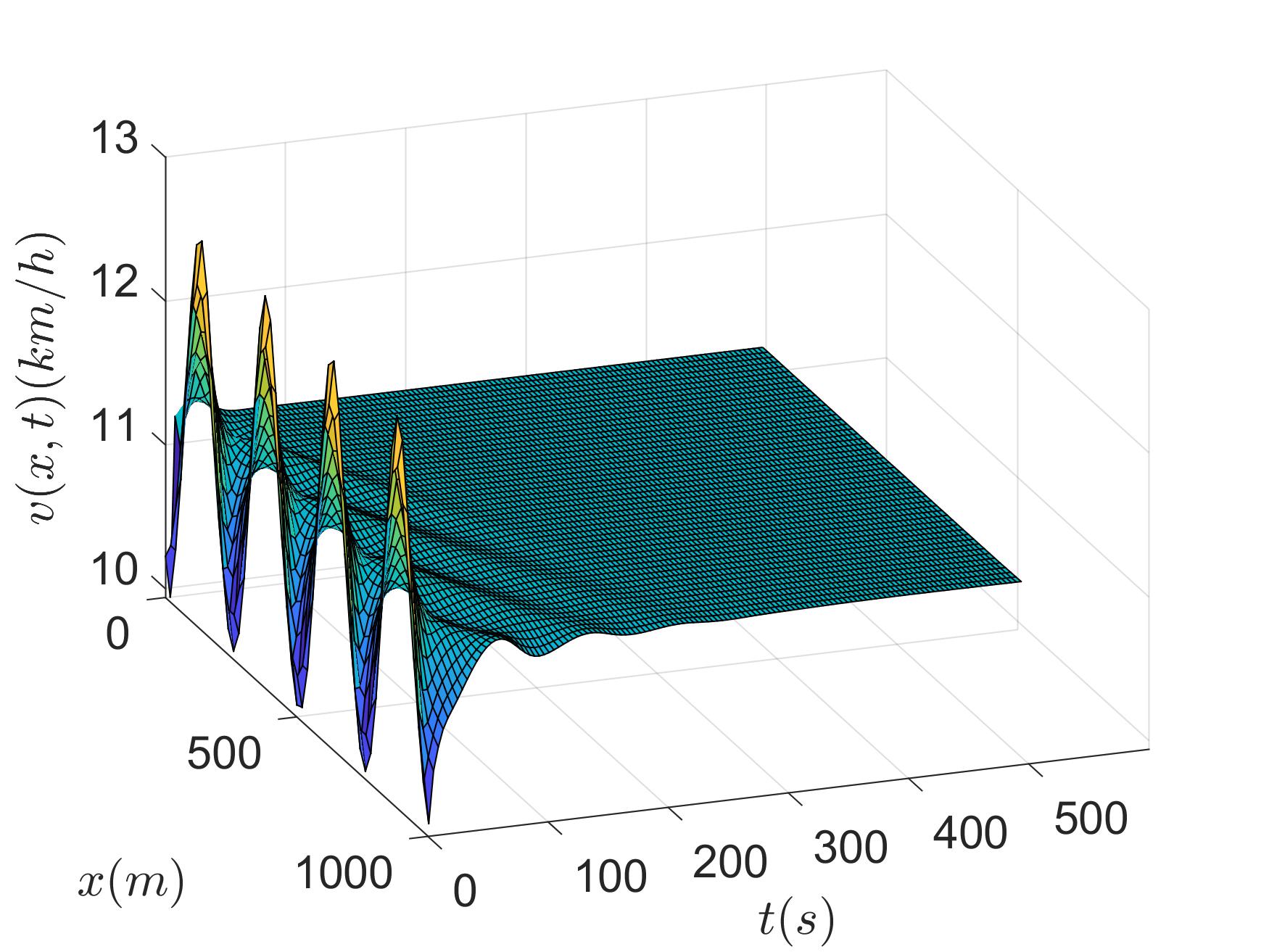}
		\end{minipage}
	}
	\subfigure[]{
		\begin{minipage}[ht]{0.25\linewidth}
			\includegraphics[width=1\linewidth]{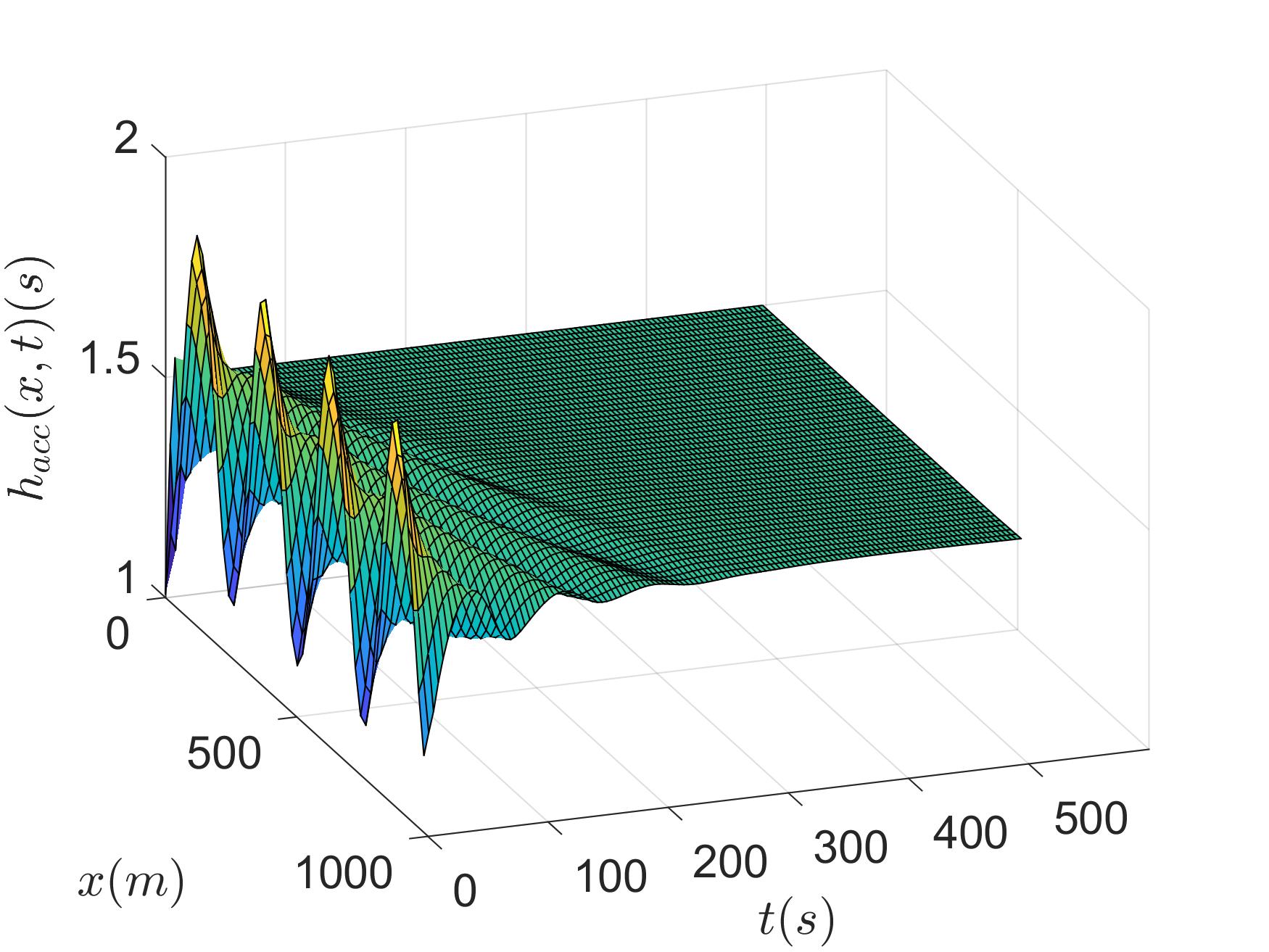}
		\end{minipage}%
	}
	\vspace{-0.3cm}
	
	\subfigure[]{
		\begin{minipage}[ht]{0.25\linewidth}
			\includegraphics[width=1\textwidth]{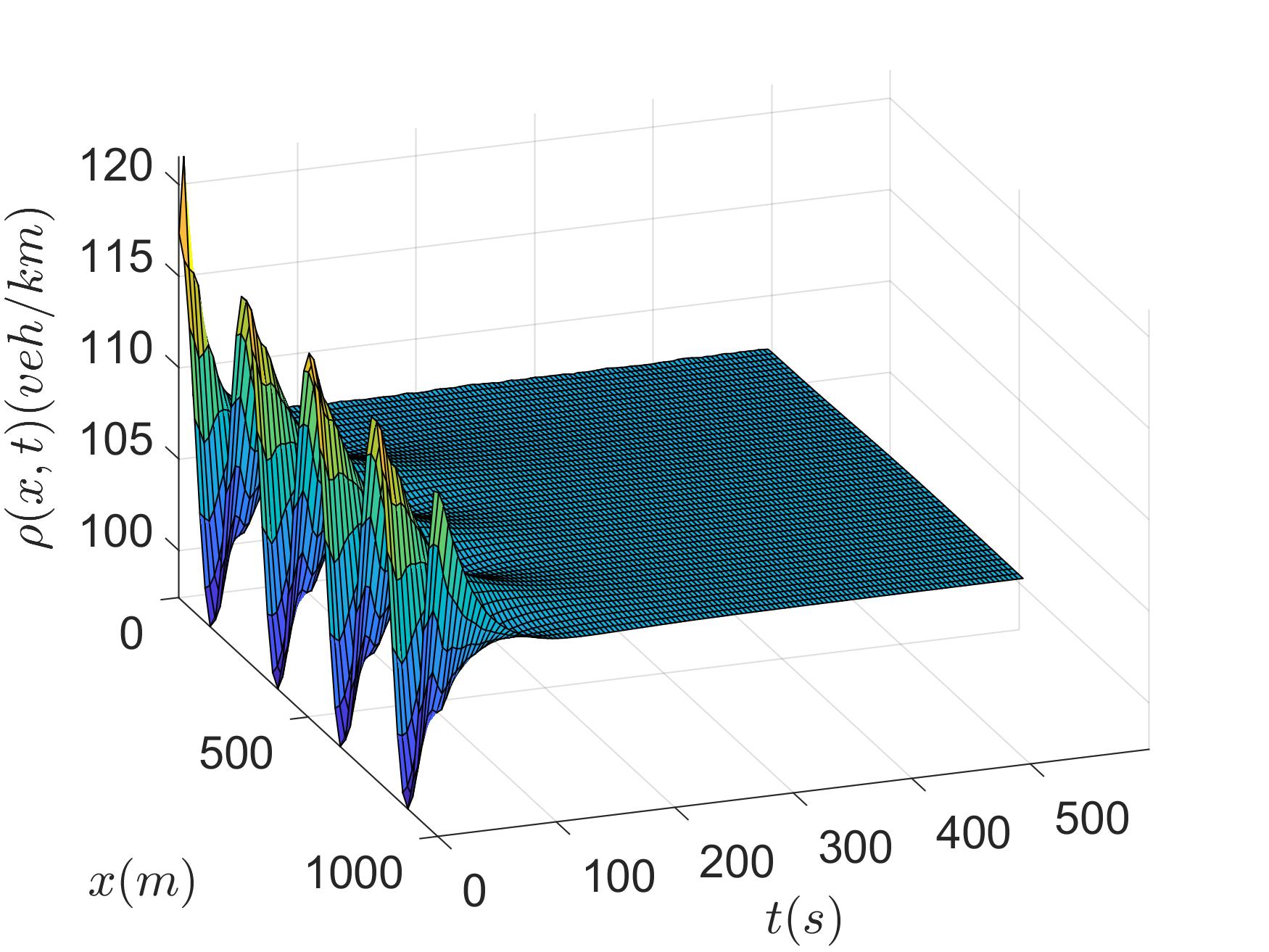}
		\end{minipage}
	}
	\subfigure[]{
		\begin{minipage}[ht]{0.25\linewidth}
			\includegraphics[width=1\linewidth]{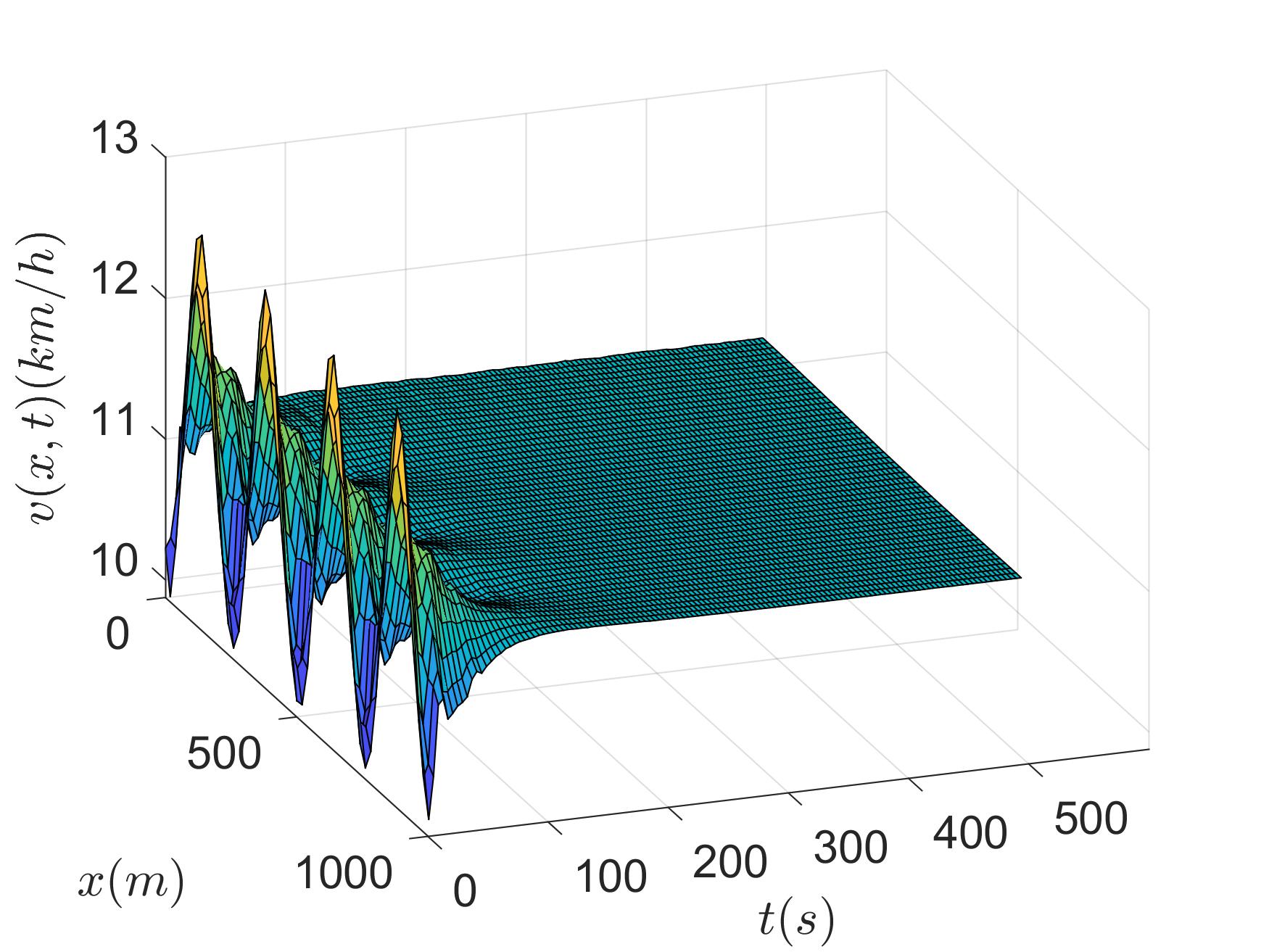}
		\end{minipage}
	}
	\subfigure[]{
		\begin{minipage}[ht]{0.25\linewidth}
			\includegraphics[width=1\linewidth]{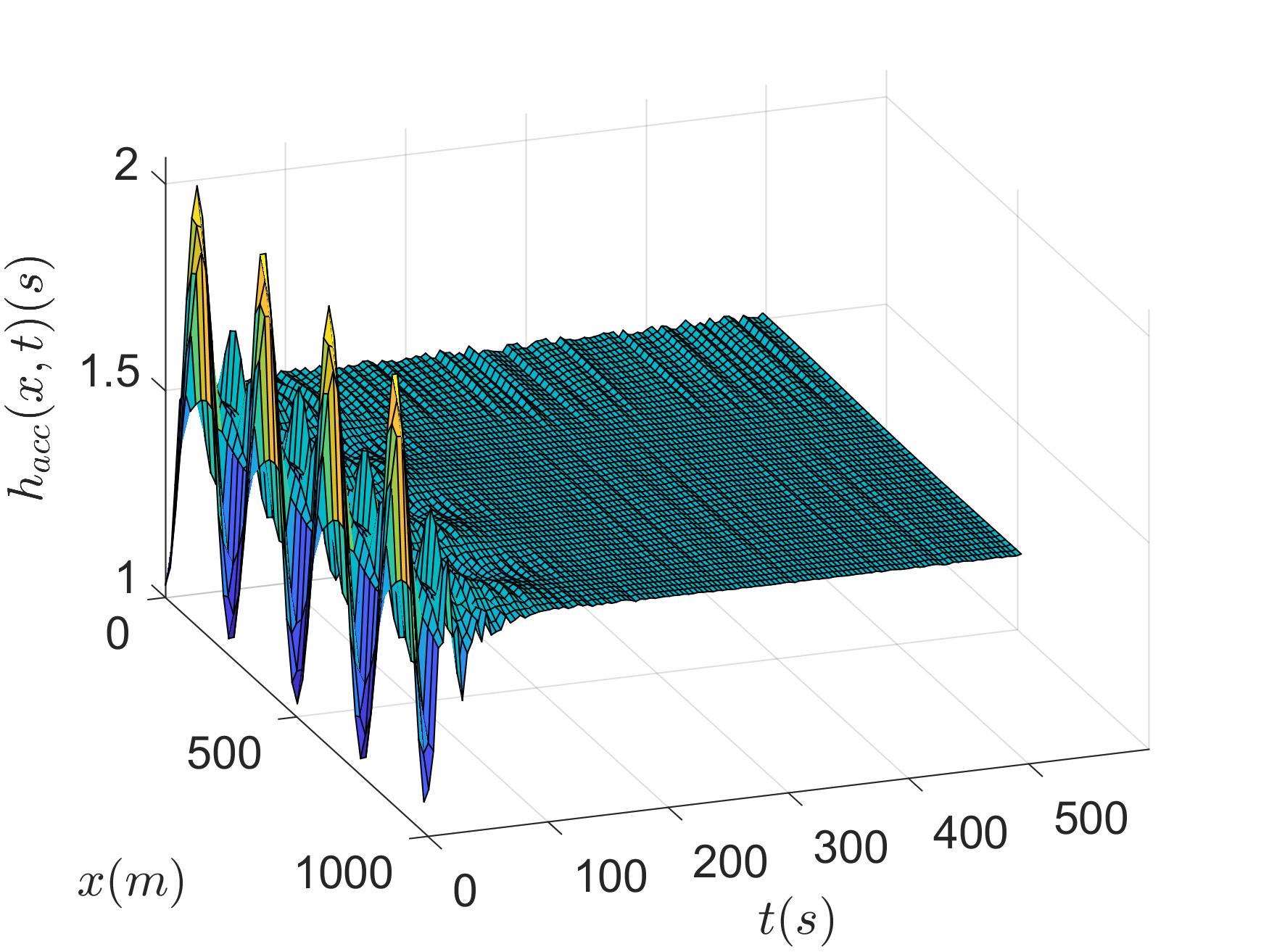}
		\end{minipage}%
	}
	\vspace{-0.3cm}
	\caption{
		(a)–(c) The states and control input of a system with a $4~s$ input delay under the Lyapunov control \cite{9018188}; (d)–(f) The states and control input of a system with a $4~s$ input delay under the backstepping control \cite{qi2021delaycompensated}; (g)–(i) The states and control input of a system with a $4~s$ input delay under the  PPO  control.
	}
	\label{fig:delay4s}
\end{figure*}

\section{Simulation results}\label{simulation}

In this section, simulation results are presented.
\subsection{Traffic Simulation}\label{traffic_simulation}
The traffic model presented in Section \ref{model} with the same parameter values as in \cite{9018188} was used in the simulations, as Table \ref{table-1} shows.
\begin{table}[h]
	\centering
	\caption{Parameters of system}\label{table-1}
	\begin{tabular}{|l|l|l|}
		\hline
		$L=1000~m$ & $l=5~m$&$q_{\rm{in}} = 1200~ veh/h$ \\
		$\alpha = 0.15$   &$\tau_{\rm{acc}}=2~s$ &$\bar{h}_{\rm{acc}}=1.5~s$
		\\$\tau_{\rm{m}} = 60~s$&
		$h_{\rm{m}} = 1~s$ & \\\hline
	\end{tabular}
\end{table}

The chosen parameters are reasonable for a traffic model\cite{belletti2015prediction,delis2015macroscopic,fan2013comparative,ngoduy2013instability,nowakowski2011cooperative}.
Let $ h_{\rm{m}} $ lower than $1.2~s$ to reflect that drivers follow a preceding vehicle at smaller time gaps in the congested traffic compared to the case of light traffic conditions.
The operating point of the traffic system, namely, the steady-state value of the mixed time gap according to \eqref{bar-h_acc}, is set to $ \bar{h}_{\rm{acc}}= 1.5 ~s $, which is a little larger than that of manual driving in heavy traffic.

The initial condition is set to $  \rho(x,0)=10\cos(8\pi x/L) $ and $  v(x,0)=q_{\rm{in}}/\rho(x,0) $ which deviates from the equilibrium value and shows a sinusoidal shape to simulate stop-and-go initial traffic conditions.

\subsection{Network Structure}
The policy-actor network and the value-critic network both have 8 layers.
For the two network, the first layer is input layer, and the other layers are hidden layers. The input layers had $600$ neurons, the second hidden layers has 1024 neurons, and the rear hidden layers contained $512$ neurons.
The outputs of the actor network are mean $\boldsymbol{\mu}^t$ and variance $\boldsymbol{\sigma}^t$ of the normal distribution that action $\boldsymbol{a}^t$ obeys. The outputs of the critic network are the state defined in \eqref{states}.

To ensure that the variance and mean are positive, the activation function of the last layer adopts the sigmoid function, and the activation function of the last critic network adopts the hyperbolic tangent function.

To avoid the variance being too small, we set the offset value to $10^{-6}$.
The learning rate is $0.001$.
The system \eqref{traffic_1}-\eqref{traffic_4} is discretized with time step $\Delta t = 0.1~s $ and spatial step $\Delta x = 5~m$. The update time is chosen as $T=10~s$, and thus $M=200$ and batch size $\epsilon=100$ equals to $N=100$.
Each batch is trained 150 times to update the parameter.
The numerical equipment is carried on CPU Intel(R) Core(TM) i9-7900X and GPU TITAN Xp.

\begin{figure}[t]
	\centering
	\includegraphics[width=0.4\linewidth]{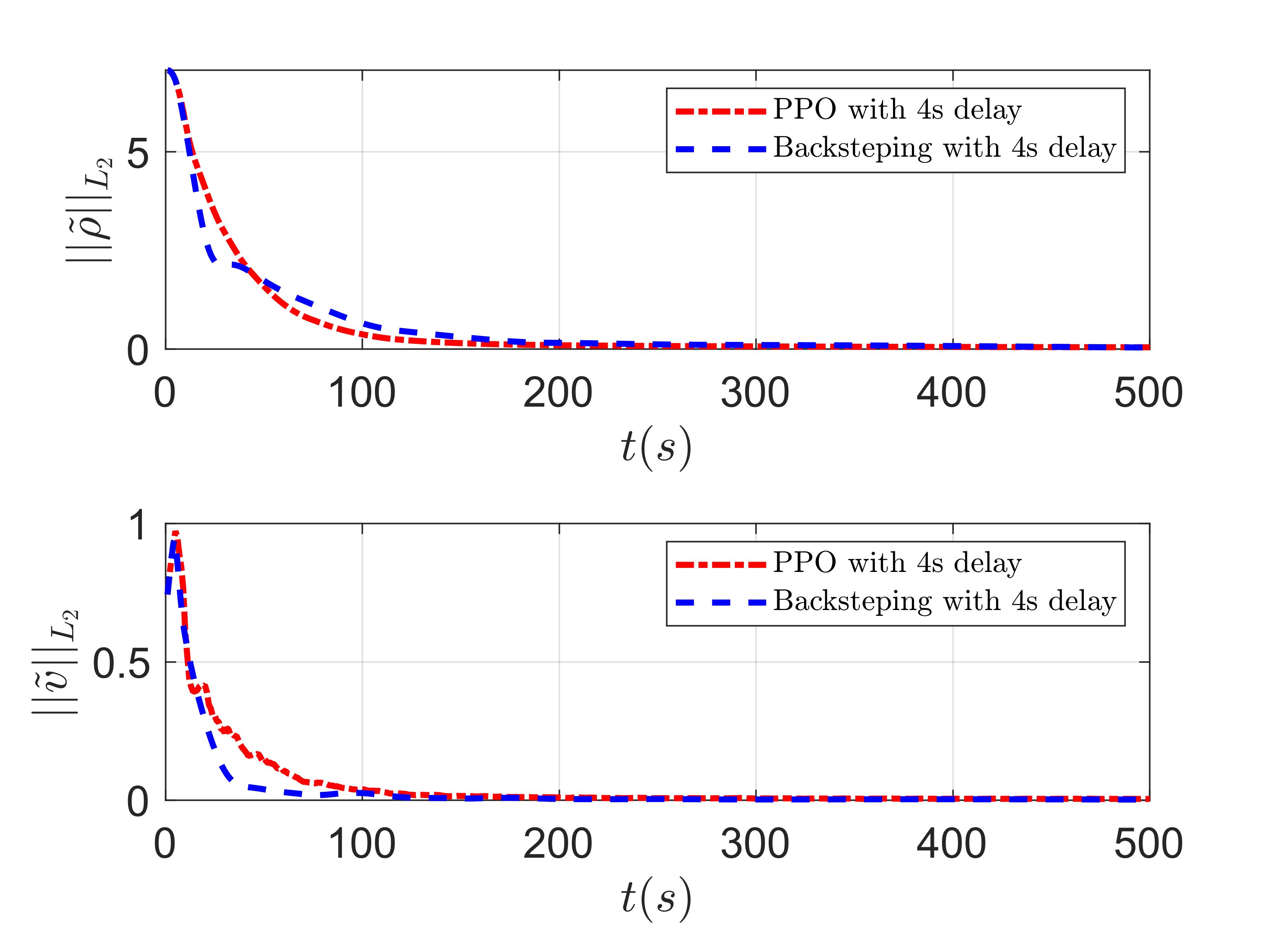}
	\caption{The $L_2$ norm of the states errors between the actual value and the desired value for a system with a $4~s$ input delay under the PPO and the backstepping control, respectively.}
	\vspace{-0.3cm}
	\label{fig:norm}
\end{figure}

\subsection{Comparison Results and Analysis}

In this subsection, the Lyapunov controller proposed in the  literature \cite{9018188} and the backstepping controller developed in \cite{qi2021delaycompensated} are used for  comparative analysis with the proposed controller.
All simulations are performed on nonlinear PDE systems.
To check the robustness of the system \eqref{traffic_1}-\eqref{traffic_4} under different controllers, we consider the two situations where a Gaussian noise is added to parameter $\alpha$ and mismatched delay value, respectively.

\begin{figure*}[t]
	\centering
	\subfigure[]{
		\begin{minipage}[ht]{0.25\linewidth}
			\centering
			\includegraphics[width=1\textwidth]{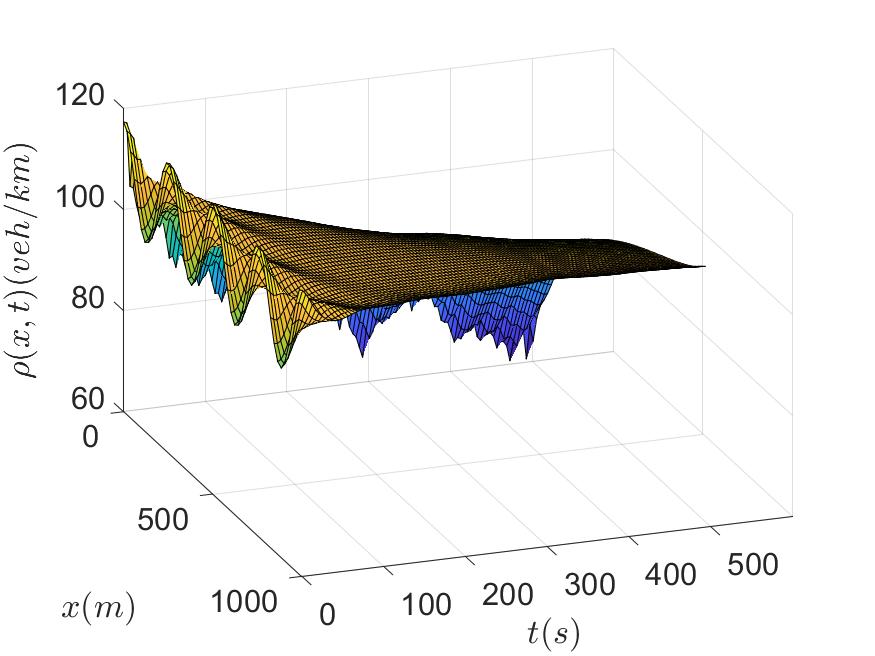}
		\end{minipage}
	}
	\subfigure[]{
		\begin{minipage}[ht]{0.25\linewidth}
			\centering
			\includegraphics[width=1\linewidth]{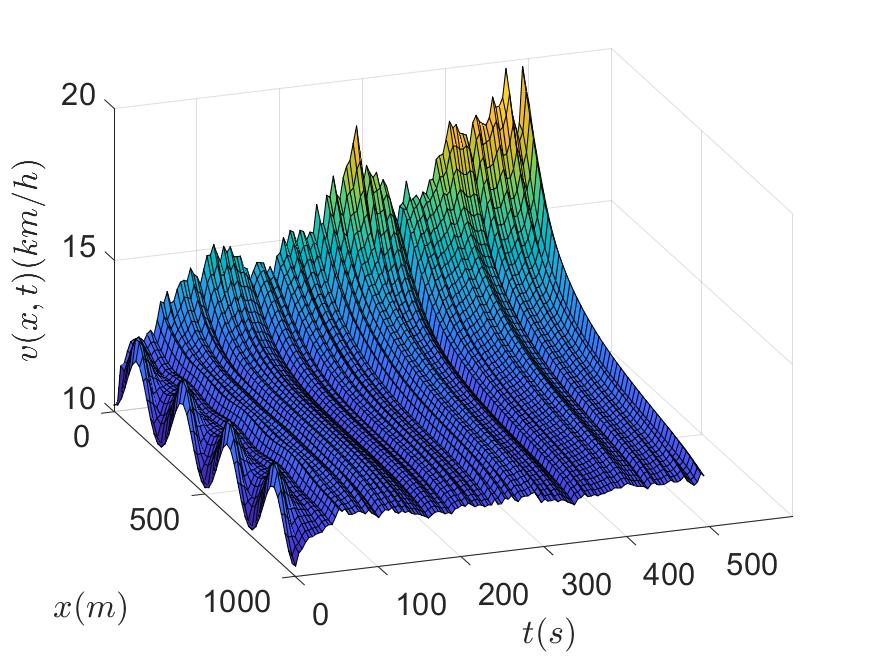}
		\end{minipage}
	}
	\subfigure[]{
		\begin{minipage}[ht]{0.25\linewidth}
			\centering
			\includegraphics[width=1\linewidth]{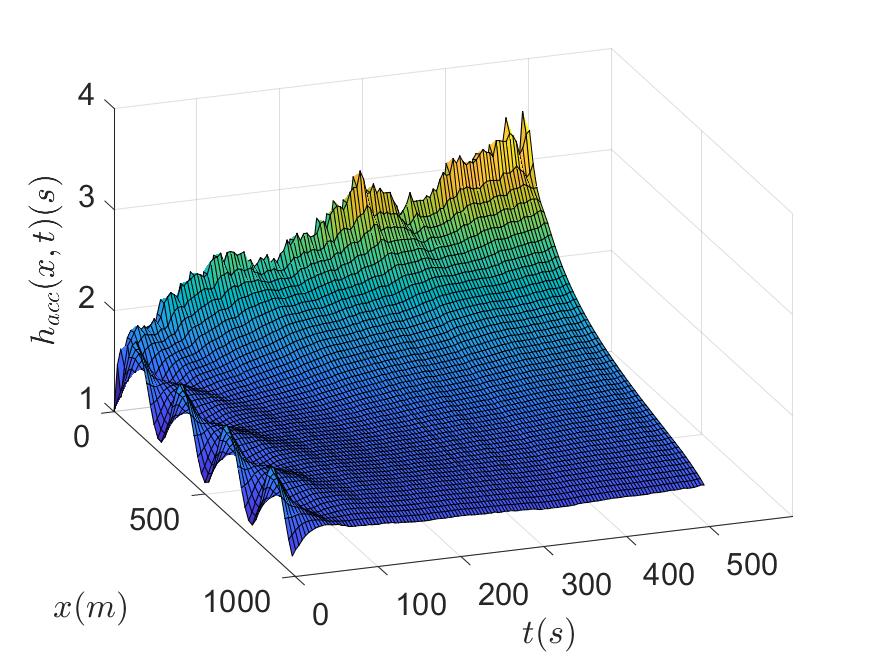}
		\end{minipage}%
	}
	\vspace{-0.3cm}
	
	\subfigure[]{
		\begin{minipage}[ht]{0.25\linewidth}
			\centering
			\includegraphics[width=1\textwidth]{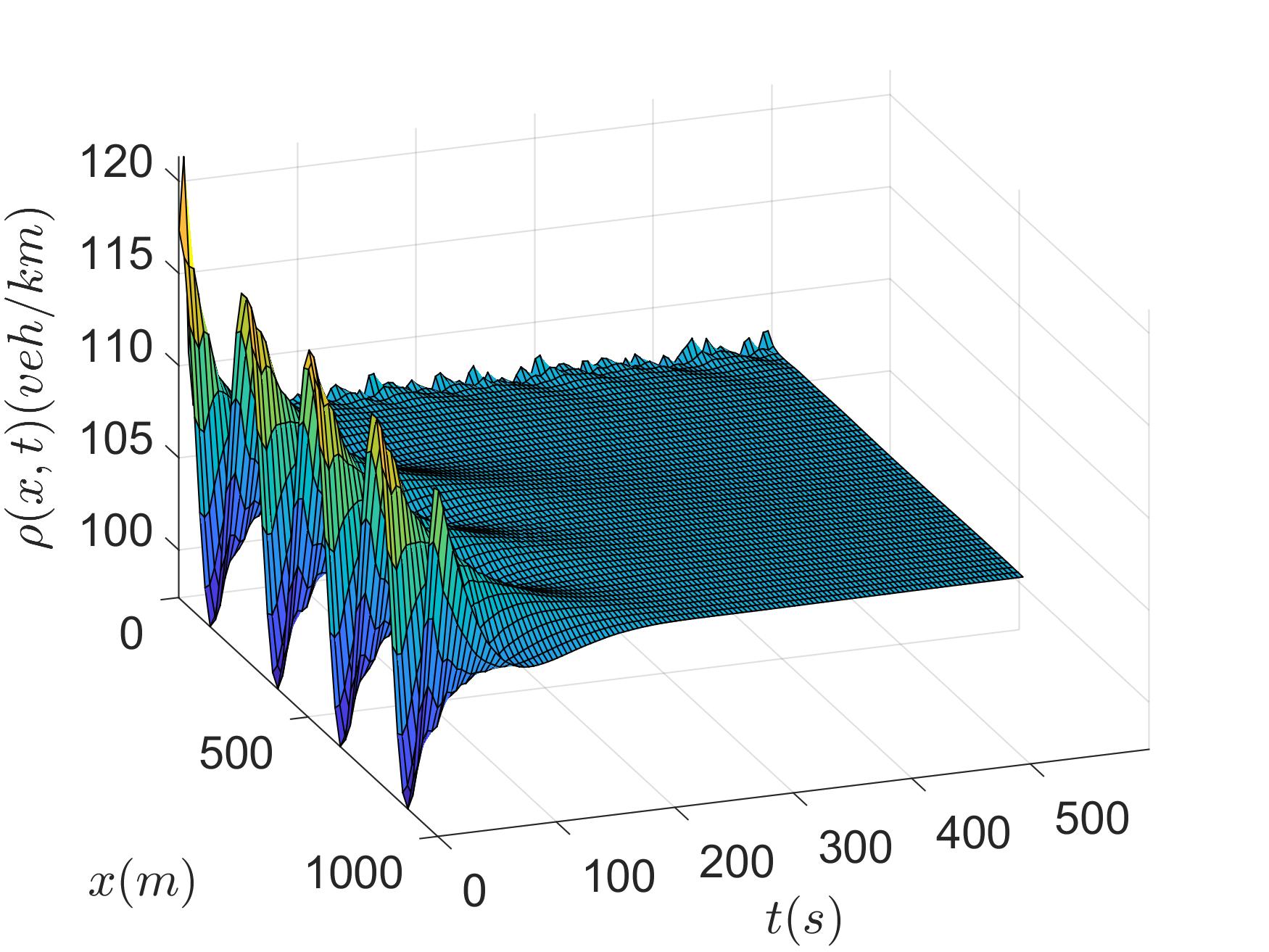}
		\end{minipage}
	}
	\subfigure[]{
		\begin{minipage}[ht]{0.25\linewidth}
			\centering
			\includegraphics[width=1\linewidth]{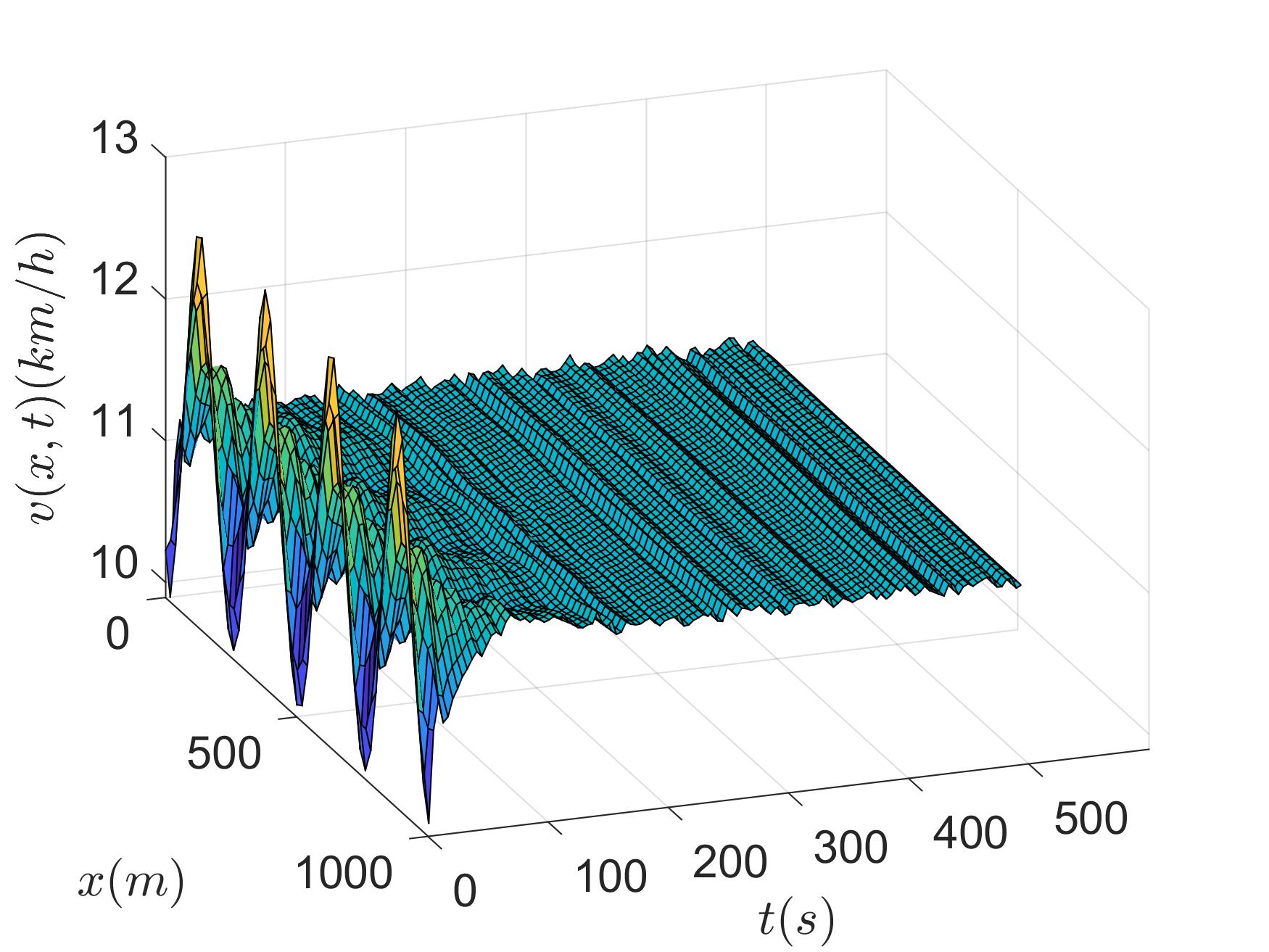}
		\end{minipage}
	}
	\subfigure[]{
		\begin{minipage}[ht]{0.25\linewidth}
			\centering
			\includegraphics[width=1\linewidth]{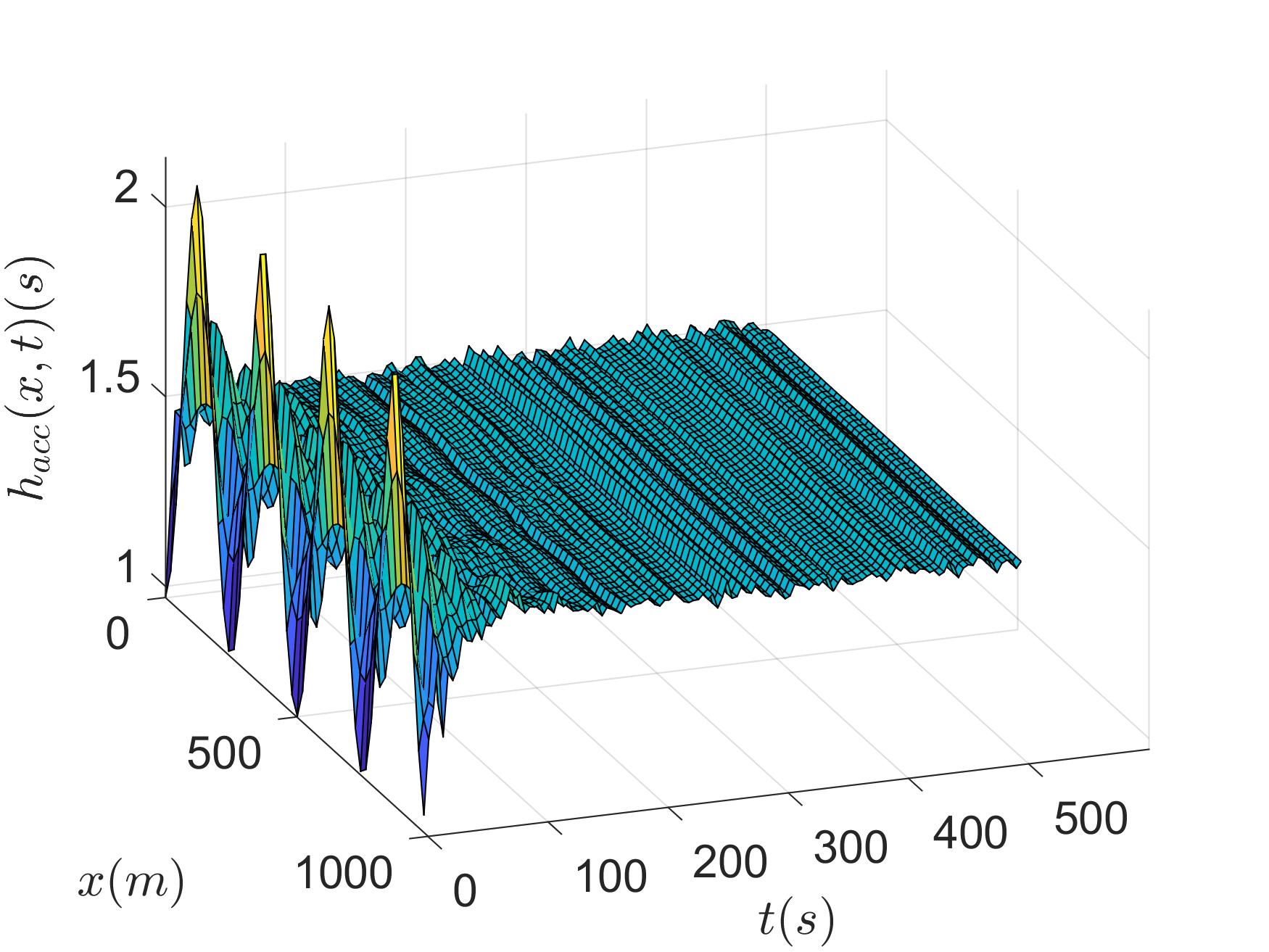}
		\end{minipage}%
	}
%
	
	\vspace{-0.3cm}
	\caption{
		(a)–(c) The states and control input of a system with a $4~s$ input delay which is subject to parameter disturbance, i.e.,  $\alpha\sim\mathcal{N}(0.15,0.15^2)$ under the backstepping control;
		(d)–(f) The states and control input of a system with a $4~s$ input delay which is subject to parameter disturbance, i.e.,  $\alpha\sim\mathcal{N}(0.15,0.15^2)$ under the PPO control;
	}
	\label{fig:tur}
\end{figure*}	

\begin{figure}[t]
	\centering
	\subfigure[]{
		\begin{minipage}[ht]{0.4\linewidth}
			\centering
			\includegraphics[width=1\textwidth]{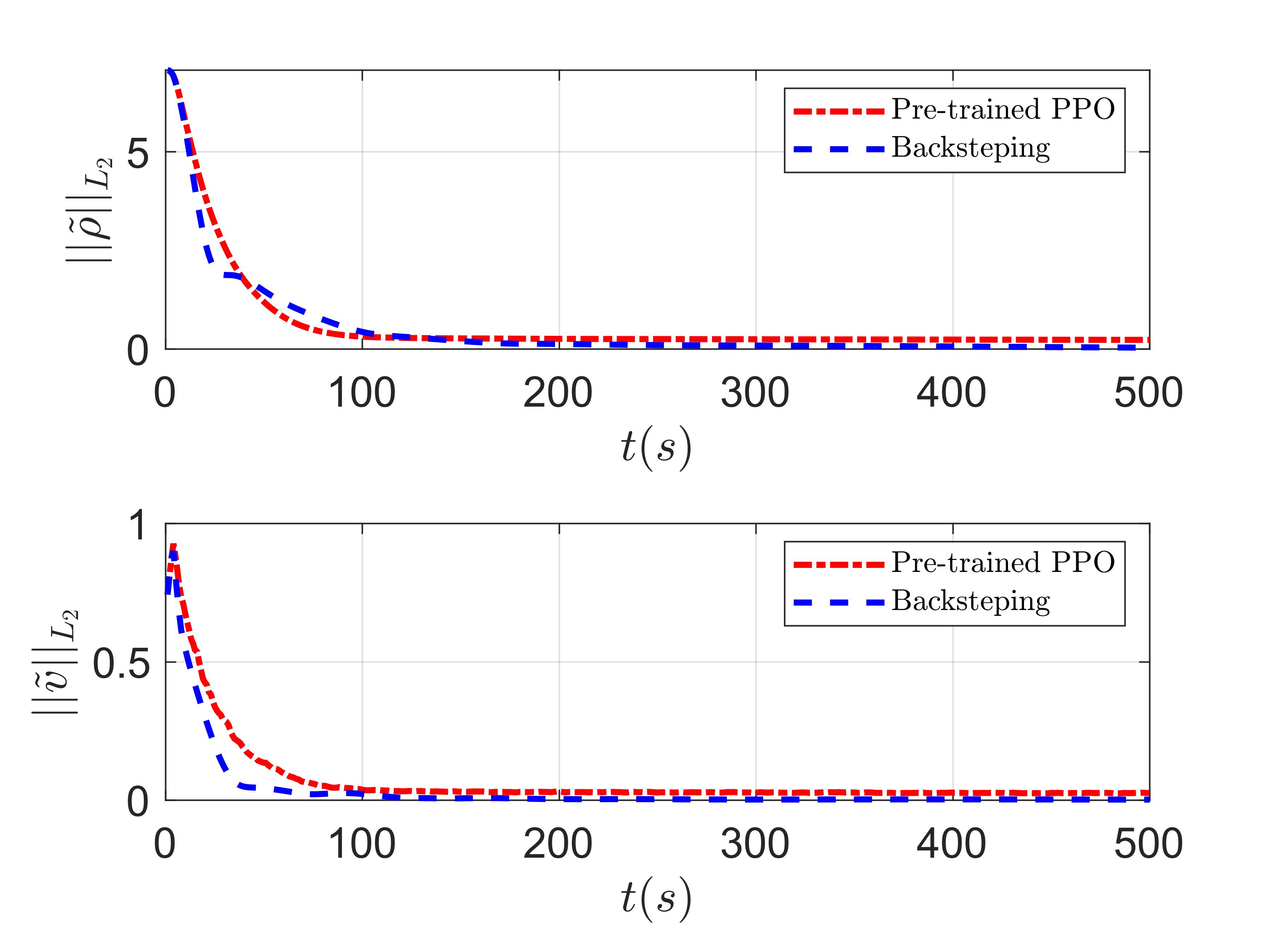}
		\end{minipage}
	}

	\subfigure[]{
		\begin{minipage}[ht]{0.4\linewidth}
			\centering
			\includegraphics[width=1\linewidth]{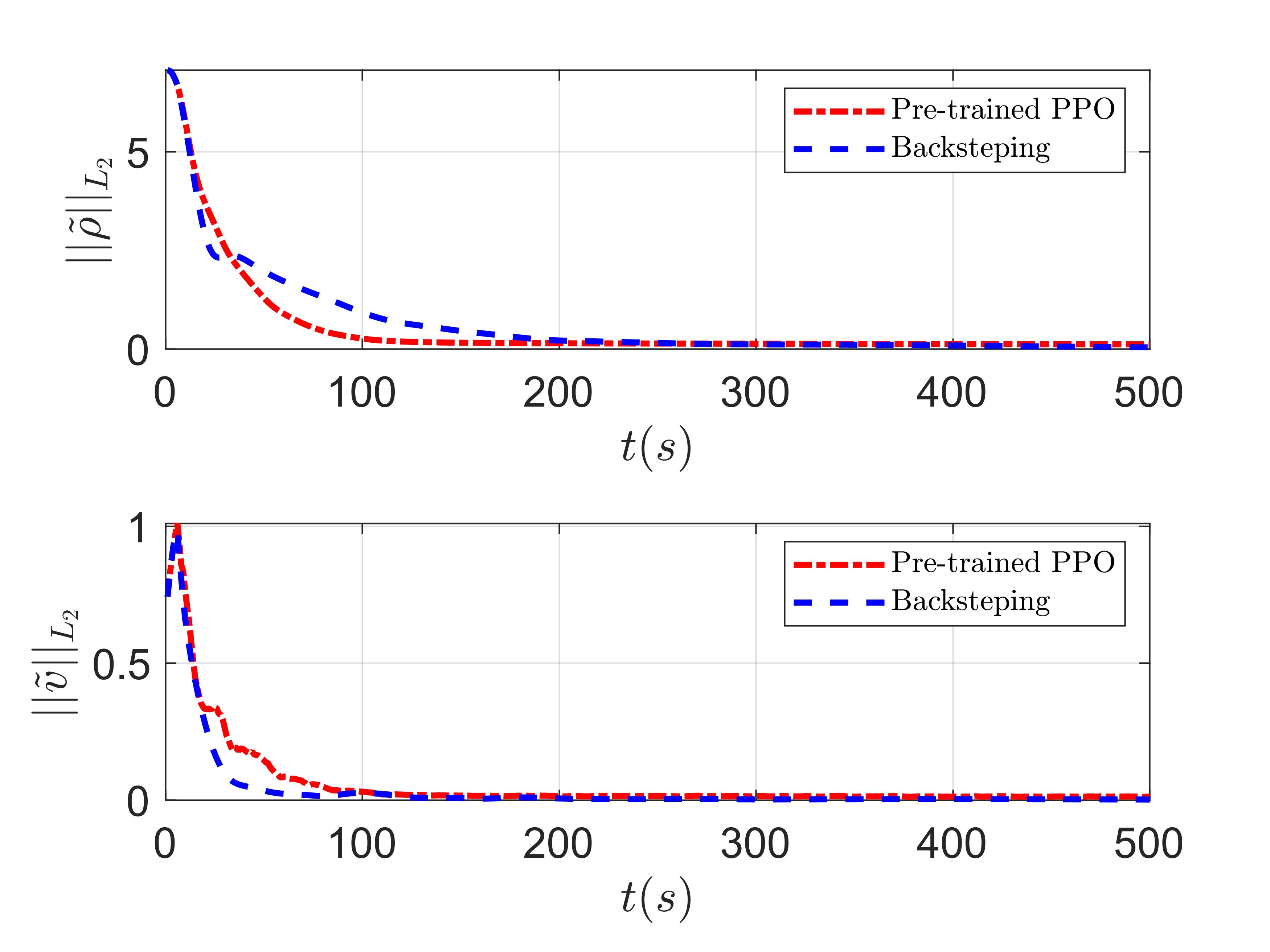}
		\end{minipage}
	}
	
	\vspace{-0.3cm}
	\caption{
		(a) The $L_2$ norm of the states errors between the actual value and the desired value of the traffic system with $4~s$ delay which is subject to parameter disturbance delay where the adopted delay $4~s$ in control is greater than the actual delay $3~s$.
		(b) The $L_2$ norm of the states errors between the actual value and the desired value of the traffic system with under-matched delay where the adopted delay $4~s$ in control is less than the actual delay $5~s$.
		}
	\label{fig:delay-unmatch}
\end{figure}	

\subsubsection{Learning process and reward}
The learning curve of the PPO  controller for the system with $4~s$ delay is shown in Figure \ref{fig:process}(a). The rewards $r$ reflecting learning performance for five different situations, is shown in  Figure \ref{fig:process}(b). To avoid random perturbations due to sampling when the reward converges the maximum, we set a stop rule for the PPO: the algorithm stops as the reward is equal to or greater than $-0.1$. As shown in Figure \ref{fig:process}(b), the system under the open-loop control cannot converge, while the reward of the PPO control converges quickly. Moreover, the PPO is robust to input delay because the reward curves of the PPO control for the system with delay and without delay almost coincide.

\subsubsection{State evolution}

The states evolution and the control input of the traffic system are shown in Figure \ref{fig:delay0s}-\ref{fig:delay4s}.
The simulation results of the traffic system are shown in Figure \ref{fig:delay0s}-\ref{fig:delay4s}. Figure \ref{fig:delay0s} illustrates the evolution of density
$\rho(x,t)$, velocity $v(x,t)$ and control input $h_{\mathrm{acc}}(x,t)$ without delay under the PPO control and Lyapunov control.
Comparing Figure \ref{fig:delay0s}(a) and \ref{fig:delay0s}(d), it is observed that the PPO controller allows the state $\rho$  converges much more quickly than the Lyapunov controller does, but as shown in Figure \ref{fig:delay0s}(b) and \ref{fig:delay0s}(e), the state $v$ converges a little faster under the Lyapunov controller than under the PPO controller.
The reason for this phenomenon is that the Lyapunov control gives control priority to velocity, which suppresses density fluctuations  by first eliminating velocity fluctuations.
However, the PPO control allows the velocity and density of the traffic flow to converge simultaneously guided by the value function.
It is observed from Figure \ref{fig:delay0s}(c) and \ref{fig:delay0s}(f) that the control effort of the PPO control is less than that of the Lyapunov control.

Figure \ref{fig:delay4s} shows the evolution of the system subject to a 4-s input delay under the Lyapunov control, the backstepping control and the PPO  control.
The results showed that the Lyapunov control cannot stabilize the $4~s$ input delayed system. The PPO  control stabilizes the system with a 4-s input delay, whose converging rate is comparable to the backstepping delay-compensator.

In order to illustrate the state evolution more clearly, we plot the $L_2$ norm of the states under the PPO and the backstepping controllers in Figure \ref{fig:norm}, with definition of the $L_2$ norm:
\begin{align}
	||\tilde \rho||_{L_{2}}=\sqrt{\frac{\sum_{i=0}^{n}(\rho(x_i,\cdot)-\bar{\rho})^2}{n}}, \\
	||\tilde v||_{L_{2}}=\sqrt{\frac{\sum_{i=0}^{n}(v(x_i,\cdot)-\bar{v})^2}{n}},
\end{align}
It can be observed from Figure \ref{fig:norm}  that the PPO control shows better convergence performance of the density compared to the backstepping control, but worse convergence performance of the velocity. The reason is also that the target system of the backstepping method is inspired by the closed-loop system with the Lyapunov control which gives control priority to velocity over density.

\subsubsection{Performance evaluation}
We apply three indexes to evaluate the performance of the closed-loop system under the different controllers. These indexes are total travel time (TTT),  fuel consumption and driving comfort defined as follows\cite{treiber2014traffic}:
\begin{align}
	J_{\mathrm{TTT}} &  = \int_{0}^{T} \int_{0}^{L} \rho(x, t) d x d t, \\
	J_{\text {fuel }} &  = \int_{0}^{T} \int_{0}^{L} \xi(x, t) \cdot \rho(x, t) d x d t,\\
	J_{\text {comfort }} & = \int_{0}^{T} \int_{0}^{L}\left(a(x, t)^{2}+a_{t}(x, t)^{2}\right) \rho(x, t) d x d t,
\end{align}
where
\begin{align}
	\xi(x, t)=&\max \{0,\nonumber\\
	 &b_{0}+b_{1} v(x, t)+b_{3} v(x, t)+b_{4} v(x, t) a(x, t)\},\\
	a(x,t) =& v_{t}(x, t)+v(x, t) v_{x}(x, t),
\end{align}
and the same values of the parameters are chosen as those in  \cite{treiber2014traffic}:
$b_0=25\cdot10^{-3}$,
$b_1=24.5\cdot10^{-6}$,
$b_3=32.5\cdot10^{-9}$,
$b_4=125\cdot10^{-6}$
and $T=300~\rm{s}$.
\begin{table*}[ht]
	\centering
	\caption{PERFORMANCE}\label{table-2}
	\setlength{\tabcolsep}{5mm}{
		\begin{tabular}{|c|c|c|c|c|c|}
			\hline
			Performance indices& Open-loop&\multicolumn{2}{c|}{Backstepping} &\multicolumn{2}{c|}{PPO} \\
			\hline
			$J_{\mathrm{TTT}}$& $3.4\times10^7$ &$3.2\times10^7$ & $5.97\%$&$3.2\times10^7$ &$5.97\%$\\\hline
			$J_{\text {fuel }}$& $8.6\times10^5$ & $8.1\times10^5$ & $5.80\%$& $8.1\times10^5$ &$5.80\%$\\\hline
			$J_{\text {comfort }}$& $9.8\times10^5$ & $1.3\times10^5$ &$87.1\%$ & $1.5\times10^5$ &$84.3\%$\\\hline
	\end{tabular}}
\end{table*}
As shown in the Table \ref{table-2}, all indexes are improved compared to the open-loop system. Only the data in the first 300 seconds are used to compute the indexes, considered that the open-loop system is unstable.
Although the PPO control has a slightly lower index of comfort than the backstepping control, it is comparable to the backstepping in other two indexes.

\subsubsection{Robustness comparison via simulation}
In the simulation, the pre-trained PPO controller is used, which is trained using the data drawn from the model with $\alpha=0.15$ and delay $D=4s$.
We perform two kinds of numerical experiments. First, we compare the robustness of the PPO control and the backstepping control to the parameter disturbance by adding a Gaussian noise $\mathcal{N}(0,0.15^2)$ to $\alpha$. Second, simulated experiments with mismatched delay values are performed, i.e., the delay value adopted by the controller is different from the actual delay.

Figure \ref{fig:tur} shows the evolution of the states considering the parameter disturbance to $\alpha$ for a system with $4~s$ input delay under the backstepping control and the PPO control. The backstepping controller fails stabilize the system with parameter disturbances, whereas the PPO controller stabilizes the system such that the state converges to the desired value, although there exist slight oscillations due to the Gaussian noise added on parameter $\alpha$.

Figure \ref{fig:delay-unmatch} shows the evolution of the state considering unmatched delay under the backstepping control and the pre-trained PPO control, respectively.
The over-matched case of the adopted delay $4~s$ in control greater than the actual delay $3~s$ is shown in Figure \ref{fig:delay-unmatch} (a), in which both controllers compensate the delay with similar convergence rate. The under-matched situation of the adopted delay $4~s$ in control less than the actual delay $5~s$ is shown in Figure \ref{fig:delay-unmatch} (b),  which illustrates the PPO controller has slight better convergence performance than  the backstepping controller.

\section{conclusion}\label{conclusion}
In this work, an PPO reinforcement learning control method is developed for stabilizing a freeway traffic flow system with input delay. The traffic flow is mixed with manual and ACC-equipped vehicles. By manipulation of the time gap of the ACC-equipped vehicles, the fluctuation of the traffic flow is suppressed which alleviates  congestion. The existing control methods heavily depend on model accuracy, while the PPO based control is model free method by training the controller through interactions with the environment that can be real system simulator from data, although in the paper, we use the ARZ model to simulate the traffic environment.

The simulation experiments illustrate that the convergence performance of the PPO control is better than that of the Lyapunov control for the delay free system. The PPO control is comparable to the backstepping control on the delay compensation for both common delay systems and mismatched delay systems although the PPO has much more simple controller form than the Backstepping. Besides, the PPO control is robust to parameter disturbance, while the backstepping control cannot stabilized the system subject to parameter disturbances.

Applying the PPO control method to the real traffic system by pre-training the networks using the real data collected from the intelligent traffic system is a promising research direction. The PPO learning algorithm can also  be improved by  modifying the reward function, resetting the state space, or changing the network parameters optimization method.

\section*{Acknowledgement}
This work was partly  supported by National Natural Science Foundation of China (62173084 and 52101346).


\begin{thebibliography}{}

\end{thebibliography}


\begin{thebibliography}{10}

\bibitem{qi2021delaycompensated}
Qi~J., Mo~S., Krstic M.. Delay-compensated distributed {PDE} control of traffic
  with connected/automated vehicles.
 {\it IEEE Trans. Autom. Control. }2022;.
 \newblock \url{https://doi.org10.1109/TAC.2022.3174032}.

\bibitem{dadashova2021multivariate}
Dadashova B., Li~X., Turner S., Koeneman P.. Multivariate time series analysis
  of traffic congestion measures in urban areas as they relate to socioeconomic
  indicators.  {\it Socio-Economic Planning Sciences. }2021;75:100877.

\bibitem{treiber2014traffic}
Treiber M., Kesting A.. Traffic Flow Dynamics: Data, Models and Simulation.
  {\it Springer-Verlag Berlin Heidelberg. }2014; 67(3): 54.

\bibitem{li2020influence}
Li~G., Lai W., Sui X., et al. Influence of traffic congestion on driver
  behavior in post-congestion driving.  {\it Accident Analysis \& Prevention.
  }2020;141:105508.

\bibitem{zhou2020modeling}
Zhou J., Zhu F.. Modeling the fundamental diagram of mixed human-driven and
  connected automated vehicles.  {\it Transportation research part C: emerging
  technologies. }2020;115:102614.

\bibitem{wen2020mapping}
Wen Y., Zhang S., Zhang J., et al. Mapping dynamic road emissions for a
  megacity by using open-access traffic congestion index data.  {\it Applied
  Energy. }2020;260:114357.

\bibitem{shen2020does}
Shen T., Hong Y., Thompson M.~M., Liu J., Huo X., Wu~L.. How does parking
  availability interplay with the land use and affect traffic congestion in
  urban areas? The case study of Xi'an, China.  {\it Sustainable Cities and
  Society. }2020;57:102126.

\bibitem{lighthill1955kinematic}
Lighthill M.~J., Whitham G.~B.. On kinematic waves II. A theory of traffic flow
  on long crowded roads.  {\it Proceedings of the Royal Society of London.
  Series A. Mathematical and Physical Sciences. }1955;229(1178):317--345.

\bibitem{richards1956shock}
Richards P.~I.. Shock waves on the highway.  {\it Operations research.
  }1956;4(1):42--51.

\bibitem{aw2000resurrection}
Aw~A., Rascle M.. Resurrection of {"second order"} models of traffic flow.
  {\it SIAM journal on applied mathematics. }2000;60(3):916--938.

\bibitem{zhang2002non}
Zhang H.~M.. A non-equilibrium traffic model devoid of gas-like behavior.  {\it
  Transportation Research Part B: Methodological. }2002;36(3):275--290.

\bibitem{fan2013data}
Fan S., Seibold B.. Data-fitted first-order traffic models and their
  second-order generalizations: Comparison by trajectory and sensor data.  {\it
  Transportation research record. }2013;2391(1):32--43.

\bibitem{kolb2017capacity}
Kolb O., G{\"o}ttlich S., Goatin P.. Capacity drop and traffic control for a
  second order traffic model.  {\it Networks \& Heterogeneous Media.
  }2017;12(4):663.

\bibitem{yu2018traffic}
Yu~H., Krstic M.. Output feedback control of two-lane traffic congestion.
 {\it  Automatica. }2021;125:109379.

\bibitem{belletti2015prediction}
Belletti F., Huo M., Litrico X., Bayen A.~M.. Prediction of traffic convective
  instability with spectral analysis of the {Aw-Rascle-Zhang} model.  {\it
  Physics Letters A. }2015;379(38):2319--2330.

\bibitem{karafyllis2018feedback}
Karafyllis I., Bekiaris-Liberis N., Papageorgiou M.. Feedback control of
  nonlinear hyperbolic {PDE} systems inspired by traffic flow models.  {\it
  IEEE Transactions on Automatic Control. }2019;64(9):3647--3662.

\bibitem{yu2019traffic}
Yu~H., Krstic M.. Traffic congestion control for {Aw-Rascle-Zhang} model.  {\it
  Automatica. }2019;100:38--51.

\bibitem{zhang2017necessary}
Zhang L., Prieur C.. Necessary and sufficient conditions on the exponential
  stability of positive hyperbolic systems.  {\it IEEE Transactions on
  Automatic Control. }2017;62(7):3610--3617.

\bibitem{Diakaki2015Overview}
Diakaki C., Papageorgiou M., Papamichail I., Nikolos I.. Overview and analysis
  of vehicle automation and communication systems from a motorway traffic
  management perspective.  {\it Transportation Research Part A: Policy and
  Practice. }2015;75:147--165.

\bibitem{9018188}
Bekiaris-Liberis N., Delis A.~I.. {PDE-based} Feedback Control of Freeway
  Traffic Flow via Time-Gap Manipulation of {ACC-equipped} Vehicles.  {\it IEEE
  Transactions on Control Systems Technology. }2021;29(1):461-469.

\bibitem{zheng2020smoothing}
Zheng Y., Wang J., Li~K.. Smoothing traffic flow via control of autonomous
  vehicles.  {\it IEEE Internet of Things Journal. }2020;7(5):3882--3896.

\bibitem{burger2019derivation}
Burger M., G{\"o}ttlich S., Jung T.. Derivation of second order traffic flow
  models with time delays.  {\it Networks \& Heterogeneous Media.
  }2019;14(2):265.

\bibitem{auriol2018delay}
Auriol J., Aarsnes U.~J.~F., Martin P., Di~Meglio F.. Delay-robust control
  design for two heterodirectional linear coupled hyperbolic {PDEs}.  {\it IEEE
  Transactions on Automatic Control. }2018;63(10):3551--3557.

\bibitem{wei2019mixed}
Wei H., Liu X., Mashayekhy L., Decker K.. Mixed-autonomy traffic control with proximal policy optimization. In 2019 IEEE Vehicular Networking Conference (VNC) (pp. 1-8). IEEE.

\bibitem{rasheed2020deep}
Rasheed F., Yau K.~A., Noor R.~M., Wu~C., Low Y.. Deep Reinforcement Learning
  for Traffic Signal Control: A Review.  {\it IEEE Access.
  }2020;8:208016-208044.

\bibitem{wang2019a}
Wang, C., Ran, B.,  Yang, H., Zhang, J and Qu, X.. A novel approach to estimate freeway traffic state: Parallel computing and improved kalman filter.
  {\it IEEE Intelligent Transportation Systems Magazine }2018;10(2):180--193.

\bibitem{yu2019reinforcement}
Yu~H., Park S., Bayen A., Moura S., Krstic M.. Reinforcement Learning Versus
  {PDE} Backstepping and {PI} Control for Congested Freeway Traffic.  {\it IEEE
  Transactions on Control Systems Technology. }2021;:1-17.

\bibitem{schulman2017proximal}
Schulman J., Wolski F., Dhariwal P., Radford A., Klimov O.. Truly proximal policy optimization. In Uncertainty in Artificial Intelligence (pp. 113-122). PMLR.

\bibitem{lillicrap2015continuous}
Shahid, A. A., Piga, D., Braghin, F.,  Roveda, L.  Continuous control actions learning and adaptation for robotic manipulation through reinforcement learning.
  {\it Autonomous Robots }2022;46(3), 483-498.


\bibitem{delis2015macroscopic}
Delis A.~I., Nikolos I.~K., Papageorgiou M.. Macroscopic traffic flow modeling
  with adaptive cruise control: Development and numerical solution.  {\it
  Computers \& Mathematics with Applications. }2015;70(8):1921--1947.

\bibitem{fan2013comparative}
Fan S., Herty M., Seibold B.. Comparative model accuracy of a data-fitted generalized aw-rascle-zhang model.  {\it Networks \& Heterogeneous Media
  }2014;9(2), 239-268.

\bibitem{ngoduy2013instability}
Ngoduy D.. Instability of cooperative adaptive cruise control traffic flow: A
  macroscopic approach.  {\it Communications in Nonlinear Science and Numerical
  Simulation. }2013;18(10):2838--2851.

\bibitem{nowakowski2011cooperative}
Song, X., Dong, Z., Ding, F., Lu, W.  Data-iteration cooperative adaptive cruise control of mixed heterogeneous vehicle platoons with unknown dynamic characteristics. {\it Asian Journal of Control. }2022;.
 \newblock \url{https://doi.org10.1002/asjc.2975}.

\end{thebibliography}

\end{document}